\documentclass[10pt,journal,compsoc]{IEEEtran}
\ifCLASSOPTIONcompsoc
  \usepackage[nocompress]{cite}
  \usepackage[linesnumbered,ruled,vlined]{algorithm2e}
  \usepackage{amsmath}
  \usepackage{amssymb}
  \usepackage{calc}
  \usepackage{color}
  \usepackage{graphicx}
  \usepackage{epstopdf}
\else
  \usepackage{cite}
\fi
\ifCLASSINFOpdf
\else
\fi
\begin{document}

\newtheorem{definition}{Definition}
\newtheorem{assumption}{Assumption}
\newtheorem{lemma}{Lemma}
\newtheorem{theorem}{Theorem}
\newtheorem{remark}{Remark}
%
\title{User-Level Privacy-Preserving Federated Learning: Analysis and Performance Optimization}
%
%
%
%

\author{Kang~Wei,~\IEEEmembership{Student Member,~IEEE,}
        Jun~Li,~\IEEEmembership{Senior Member,~IEEE,}
        Ming~Ding,~\IEEEmembership{Senior Member,~IEEE,}
        Chuan~Ma,
        Hang~Su,~\IEEEmembership{Senior Member,~IEEE,}
        Bo~Zhang,~\IEEEmembership{Senior Member,~IEEE,}
        and~H.~Vincent~Poor,~\IEEEmembership{Life~Fellow,~IEEE}
\IEEEcompsocitemizethanks{\IEEEcompsocthanksitem Kang~Wei, Jun~Li (corresponding author) and Chuan~Ma are with School of Electrical and Optical Engineering, Nanjing University of Science and Technology, Nanjing, China. E-mail: \{kang.wei, jun.li, chuan.ma\}@njust.edu.cn.
\IEEEcompsocthanksitem Ming~Ding is with Data61, CSIRO, Sydney, Australia. E-mail: ming.ding@data61.csiro.au.
\IEEEcompsocthanksitem Hang~Su and Bo~Zhang are with Department of Computer Science and Technology, Tsinghua University, Beijing, China. E-mail: \{suhangss, dcszb\}@tsinghua.edu.cn.
\IEEEcompsocthanksitem H.~Vincent~Poor is with Department of Electrical Engineering, Princeton University, NJ, USA. E-mail: poor@princeton.edu.}
\thanks{Manuscript received April 19, 2005; revised August 26, 2015.}}

%
%

\markboth{Journal of \LaTeX\ Class Files,~Vol.~14, No.~8, August~2015}%
{Shell \MakeLowercase{\textit{et al.}}: Bare Demo of IEEEtran.cls for Computer Society Journals}
%



\IEEEtitleabstractindextext{%
\begin{abstract}
Federated learning (FL), as a type of collaborative machine learning framework, is capable of preserving private data from mobile terminals (MTs) while training the data into useful models. Nevertheless, from a viewpoint of information theory, it is still possible for a curious server to infer private information from the shared models uploaded by MTs. To address this problem, we first make use of the concept of local differential privacy (LDP), and propose a user-level differential privacy (UDP) algorithm by adding artificial noise to the shared models before uploading them to servers. According to our analysis, the UDP framework can realize $(\epsilon_{i}, \delta_{i})$-LDP for the $i$-th MT with adjustable privacy protection levels by varying the variances of the artificial noise processes. We then derive a theoretical convergence upper-bound for the UDP algorithm. It reveals that there exists an optimal number of communication rounds to achieve the best learning performance. More importantly, we propose a communication rounds discounting (CRD) method. Compared with the heuristic search method, the proposed CRD method can achieve a much better trade-off between the computational complexity of searching and the convergence performance. Extensive experiments indicate that our UDP algorithm using the proposed CRD method can effectively improve both the training efficiency and model quality for the given privacy protection levels.
\end{abstract}

\begin{IEEEkeywords}
Federated learning, differential privacy, communication round, mobile edge computing.
\end{IEEEkeywords}
}

\maketitle

\IEEEdisplaynontitleabstractindextext

%
\IEEEpeerreviewmaketitle

\IEEEraisesectionheading{\section{Introduction}\label{sec:introd}}

%
%
%
%
\IEEEPARstart{W}{ith} the dramatic development of the internet-of-things (IoT), the amount of data originating from intelligent devices is growing at unprecedented rates~\cite{Li2015Distributed,Deng2020Wireless}.
Conventional machine learning (ML) is no longer capable of efficiently processing such data in a centralized manner.
To address this challenge, several distributed ML architectures have been proposed with different approaches of aggregating gradients or models~\cite{Abadi2016Deep}.
However, data privacy and confidentiality~\cite{fang2013public,S2020Privacy,Ge2020Secure} are of concern in such approaches as exchanged gradients or models usually contain clients' sensitive information.
One such distributed ML architecture is federated learning (FL), which allows a decoupling of data provision at end-user equipment (UE) and machine learning model aggregation at a central server~\cite{McMahan2016Fed,Jakub2016Fed}.
In FL, all mobile terminals (MTs) with the same data structure collaboratively learn a shared model with the help of a server, i.e., training the model at MTs and aggregating model parameters at the server. Owing to the local training, FL does not require MTs to upload their private data, thereby effectively reducing transmission overhead as well as promoting MTs' privacy. As such, FL is applicable to a variety of scenarios where data are either sensitive or expensive to be transmitted to the server, e.g., health-care records, private images, and personally identifiable information, etc.~\cite{Ma2019FL,Jiao2020Toward,Nguyen2020Enabling}.

Although FL can help prevent private data from being exposed to the public, hidden adversaries may attack the learning model by eavesdropping and analyzing the shared parameters, e.g., via a reconstruction attack~\cite{Wang2019Beyond} or an inference attack~\cite{Melis2019Exploiting}. For example, a malicious classifier may reveal the features of the MTs' data and reconstruct data points from the FL training process~\cite{Wang2019Beyond}. Some designed attack strategies can be found in recent studies. The work in~\cite{Phong2017Privacy} recovered the private data based on the observation that the gradient of weights is proportional to that of the bias, and their ratio approximates the training input. The work in~\cite{Wang2019Beyond} considered an untrusted server in FL and proposed a generative adversarial network (GAN) based reconstruction attack. Furthermore, this reconstruction attack utilized the shared model as the discriminator to train a GAN~\cite{Goodfellow2014GAN} model that generates original samples of the training data. In~\cite{Melis2019Exploiting}, Melis~\emph{et al.} demonstrated that the shared models in FL may leak unintended information from participants' training data, and they developed passive and active inference attacks to exploit this leakage.
The work in~\cite{Zhu2019Deep} showed that it is possible to obtain the private training data from the publicly shared gradients, i.e., deep leakage from gradient, which was empirically validated on both computer vision and natural language processing tasks.
The work in~\cite{Wang2019Beyond} proposed a novel GAN framework with a multi-task discriminator at the server side to attack user-level privacy, which can simultaneously discriminate category, reality, and user identity of input samples.
Therefore, it is challenging to preserve data contributors' privacy from~\cite{ma2020rdpgan}.

Therefore, privacy-preserving ML has attracted intensive attention in recent years, as the emergence of centralized searchable data repositories and open data applications may lead to the leakage of private information.
The work in~\cite{Abadi2016Deep} first proposed the concept of deep learning with DP, providing an evaluation criterion for privacy guarantees.
The work in~\cite{Lee2018Concentrated} improved DP based stochastic gradient descent (SGD) algorithms by carefully allocation of a privacy budget at each training iteration.
In~\cite{Lee2018Concentrated}, the privacy budget and step size for each iteration are dynamically determined at runtime based on the quality of the noisy statistics (e.g., gradient) obtained for the current training iteration.
Privacy issues are also more critical in distributed ML.
The work in~\cite{Li2018Differentially} introduced the notion of DP in distributed ML and proposed a distributed online learning algorithm to improve the learning performance for a given privacy level.
The work in~\cite{Huang2020ADMM} analyzed the privacy loss in a DP-based distributed ML framework, and provided the explicit convergence performance.
The work in~\cite{Wu2019The} presented a theoretical analysis of DP-based SGD algorithms, and provided an approach for analyzing the quality of ML models related to the privacy level and the size of the datasets.

A formal treatment of privacy risks in FL calls for a holistic and interdisciplinary approach~\cite{Wei2020Fed}.
The work in~~\cite{Robin2017Differentially} proposed an FL algorithm based on a random sub-sampling scheduling at each aggregation.
This algorithm can achieve good training performance at a given privacy level when there are a sufficiently large number of participating MTs.
However, it cannot preserve MTs' private information from being exposed to a curious server.
The work in~\cite{Truex2019HAP} presented an alternative approach that utilizes both DP and secure multiparty computation (SMC) to protect against inference threats and produces models with high accuracy.
A key component of this work is the ability to reduce noise by leveraging the SMC framework while considering a customizable trust parameter, which will also consume more communication and computing resources.
The work in~\cite{li2019privacy} involved sketching algorithms, using hash functions to compress the input data with bounded errors, to consider communication efficiency and privacy protection in distributed learning.
However, some of these approaches cannot preserve MTs' private information from being exposed to the curious server.
Moreover, other approaches, such as SMC and sketching algorithms, can consume large amounts of communications and computer resources.

To solve these challenges, it is important to design an effective algorithm to mitigate the privacy concerns for data sharing without deteriorating the quality of trained FL models. In this paper, in order to prevent information leakage from the shared model parameters and improve the training efficiency, we develop a novel privacy-preserving FL framework, termed the user-level differential privacy (UDP) algorithm. Furthermore, we develop a theoretical convergence bound for this UDP algorithm and design a novel online optimization method to achieve better convergence performance compared with the original UDP algorithm.

Specifically, the contributions of this paper can be summarized as follows:
\begin{itemize}
\item[$\bullet$] We introduce a novel privacy-preserving FL framework, namely the user-level differential privacy (UDP) algorithm, using the concept of local differential privacy (LDP) and verify that this UDP framework can realize $(\epsilon_{i}, \delta_{i})$-LDP for the $i$-th MT for iterative learning model exchange with a curious server based on our analysis.
\item[$\bullet$] We enhance a standard differential privacy (DP) mechanism (the moments accountant method) and analyze the sensitivity for each MT under the LDP definition. Then, we prove that the training process of the $i$-th MT satisfies the requirement of $(\epsilon_{i}, \delta_{i})$-LDP for different privacy levels by properly adapting their variances of Gaussian noise added to the model updates.
\item[$\bullet$] We also show that there exists an optimal number of communication rounds in terms of convergence performance for a given privacy level. This property demonstrates that we need to have a new look at communication rounds. Thus, we design an online optimization method, termed communication rounds discounting (CRD), which can obtain a better tradeoff between complexity and convergence performance compared with the original UDP algorithm and an offline heuristic search method.
\item[$\bullet$] We conduct extensive experiments on real-word datasets and the experimental results validate that our CRD method in UDP can achieve performance equivalent to that of offline search in terms of loss function values, but with a much lower complexity.
\end{itemize}

The remainder of this paper is organized as follows.
The threat model and background on LDP and FL are presented in Section~\ref{sec:Preliminaries}.
Then, we provide the details of the proposed UDP algorithm and the privacy analysis in Section~\ref{sec:Con_FL},
and present the noise recalculation and CRD methods in Section~\ref{Sec:dyn_alo}.
The analytical and experimental results are shown in Section~\ref{Sec:Exm_Res}.
Finally, conclusions are drawn in Section~\ref{Sec:Concl}.

\begin{table}[ht]\caption{Summary of Main Notation}
\centering
\begin{tabular}{c|c}
\hline
$\mathcal M$& A randomized mechanism for LDP\\
\hline
$x,x'$& Adjacent databases\\
\hline
$\epsilon, \delta$& The parameters related to the original LDP\\
\hline
$\epsilon_{i}, \delta_{i}$& The parameters related to LDP for the $i$-th MT\\
\hline
$\mathcal C_i$& The $i$-th MT\\
\hline
$\mathcal D_i$& The dataset held by the owner $\mathcal C_i$\\
\hline
$\mathcal D_{i,m}$& The $m$-th sample in $\mathcal D_i$\\
\hline
$\mathcal D$& The dataset held by all the MTs\\
\hline
$|\cdot|$& The cardinality of a set\\
\hline
$\mathcal{U}$& The set of all users\\
\hline
$U$& Total number of all users ($U=\vert\mathcal{U}\vert$)\\
\hline
$\mathcal{K}$& The set of chosen MTs\\
\hline
$K$& The number of chosen MTs ($K=\vert\mathcal{K}\vert$)\\
\hline
$t$& The index of the $t$-th communication round\\
\hline
$T$& The number of communication rounds\\
\hline
$\boldsymbol{w}$& The vector of model parameters\\
\hline
$F(\boldsymbol{w})$& Global loss function\\
\hline
$F_{i}(\boldsymbol{w})$& Local loss function from the $i$-th MT\\
\hline
$\boldsymbol{w}_{i}^{t}$& Local training parameters of the $i$-th MT\\
\hline
$\boldsymbol{\widetilde{w}}_{i}^{t}$& Local training parameters after adding noises\\
\hline
$\boldsymbol{w}^{0}$& Initial parameters of the global model\\
\hline
$\boldsymbol{w}^{t}$& Global parameters generated from local\\
\hline
&parameters at the $t$-th communication round\\
\hline
$\boldsymbol{w}^{*}$& The optimal parameters that minimize $F(\boldsymbol{w})$\\
\hline
\end{tabular}
\label{tab:SummaryofMainNotations}
\end{table}
\section{Preliminaries}\label{sec:Preliminaries}
In this section, we first present the FL framework and the threat model, and then introduce the basic knowledge of $(\epsilon,\delta)$-LDP.
\subsection{Federated Learning}
Let us consider a general FL system consisting of a centralized server and $U$ MTs.
Let $\mathcal D_i$ denote the local dataset held by MT $\mathcal C_i$, where $\mathcal{U} = \{1, 2,\ldots, U\}$ and  $i\in \mathcal{U}$. For all participants, the objective is to learn a global model over data that resides at the $U$ associated MTs.
Formally, this FL task can be expressed as
\begin{equation}\label{equ:global_loss}
\boldsymbol{w}^{*}=\mathop{\arg\min}_{\boldsymbol{w}}{F(\mathcal D,\boldsymbol{w})},
\end{equation}
where $F(\mathcal D,\boldsymbol{w})=\sum_{i\in \mathcal{U}}p_{i}F_{i}(\mathcal D_{i},\boldsymbol{w})$, $F_{i}(\cdot)$ is the local loss function of the $i$-th MT, $p_{i} = \vert \mathcal D_i\vert/\vert \mathcal D\vert\geq 0$ with $\sum_{i\in \mathcal{U}}{p_{i}}=1$, $\vert \mathcal D_{i}\vert $ is the number of data samples in the $i$-th MT's dataset and $\vert \mathcal D\vert = \sum_{i\in \mathcal{U}}{\vert \mathcal D_{i}\vert}$ is the total number of data samples in all MTs' datasets, respectively. Generally, the local loss functions $F_{i}(\cdot)$ are given by local empirical risks and have the same expression for various MTs.
At the server, $K$ MTs are chosen and a model aggregation is performed over their uploaded models. In particular, $\boldsymbol{w}$ is the global model parameter, given by
\begin{equation}\label{equ:aggregation}
\boldsymbol{w} = \sum_{i\in \mathcal{K}}{p_{i}\boldsymbol{w}_{i}},\tag{2}
\end{equation}
where $\boldsymbol{w}_{i}$ is the parameter vector trained at the $i$-th MT, $\mathcal{K}$ is a subset of $\mathcal{U}$, with $K$ MTs out of $U$ MTs chosen for participating in the model aggregation, respectively.
To facilitate analysis on the privacy performance, in this paper, we adopt a low complexity method, termed $K$-user random scheduling, in which $K$ MTs are randomly chosen from all the MTs.

\subsection{Threat Model}
In this paper, we assume that the server is honest-but-curious, which means it will strictly follow the FL rule, but may recover the training datasets~\cite{Wang2019Beyond} or infer private features~\cite{Melis2019Exploiting} based on the local uploaded parameters.
Concretely, this curious server can train a GAN framework, e.g., multi-task GAN-AI~\cite{Wang2019Beyond}, which may simultaneously discriminates category, reality, and user identity of input samples.
This novel discrimination on user identity enables the generator to recover users' private data.
In addition, this curious server may also be interested in learning whether a given sample belongs to the training datasets, which can be inferred by utilizing the difference between model outputs from training and non-training this sample~\cite{Shokri2017Membership}.
For example, when FL is conducting a clinical experiment, a participant may not want the observer to know whether he is involved in this experiment (a MT may contain several individuals' records).
However, the adversary may link the test results to the appearance or disappearance of this participant, and then possibly inflict harm to this person.
Therefore, our threat model is reasonable and realistic.
\subsection{Local Differential Privacy}
DP mechanism with parameters $\epsilon$ and $\delta$ provides a strong criterion for the privacy preservation of distributed data processing systems.
Here, $\epsilon > 0$ is the distinguishable bound of all outputs on neighboring datasets $x, x'$ in a database $\mathcal{X}$, and $\delta$ represents the probability of the event that the ratio of the probabilities for two adjacent datasets $x, x'$ cannot be bounded by $e^{\epsilon}$ after adding a privacy-preserving mechanism. With an arbitrarily given $\delta$, a larger $\epsilon$ gives a clearer distinguishability of neighboring datasets and hence a higher risk of privacy violation.
Now, we will formally define LDP as follows.
\begin{definition}($(\epsilon, \delta)$-LDP~\cite{Wang2019Local}):
A randomized mechanism $\mathcal M$ satisfies $(\epsilon, \delta)$-DP: $\mathcal{X}\rightarrow \mathcal{R}$ with domain $\mathcal{X}$ and range $\mathcal{R}$, for all measurable sets $\mathcal S\subseteq \mathcal{R}$ and any two adjacent datasets $x, x'\in \mathcal{X}$, we have
\begin{equation}
\emph{Pr}[\mathcal M(x)\in \mathcal S]\leq e^{\epsilon}\emph{Pr}[\mathcal M(x')\in \mathcal S]+\delta.
\end{equation}
\end{definition}

In this paper, we choose the Gaussian mechanism that adopts $L_2$ norm sensitivity.
It adds zero-mean Gaussian noise with variance $\sigma^2\mathbf{I}$ in each coordinate of the function output $s(x)$ as
\begin{equation}
\mathcal M(x) = s(x)+\mathcal N(0, \sigma^2\mathbf{I}),
\end{equation}
where $\mathbf{I}$ is an identity matrix and has the same size with $s(x)$.
The sensitivity of the function $s$ can be expressed as
\begin{equation}\label{eq:org_sensi}
\Delta s = \max_{x, x'\in \mathcal{X}}\Vert s(x)-s(x')\Vert_{2},
\end{equation}
which gives an upper bound on how much we must perturb its output considering preserving privacy.
It satisfies $(\epsilon, \delta)$-LDP when we properly select the value of $\sigma$.
\section{Privacy Analysis}\label{sec:Con_FL}
In this section, we first propose a user-level differential privacy (UDP) algorithm in FL against the curious server.
Then, we develop a method for analyzing the privacy loss moment by improving the conventional analysis approach~\cite{Abadi2016Deep}.
Based on our improved method, we are able to accurately calculate the standard deviation (STD) of additive noises in UDP to guarantee $(\epsilon_{i}, \delta_{i})$-LDP for the $i$-th MT.
\subsection{User-level DP}
In this subsection, to prevent information leakage from uploaded parameters, we will introduce a UDP algorithm, which borrows the concept of LDP.
Our UDP algorithm is outlined in~\textbf{Algorithm~\ref{alg:UDP}} for training an effective model with the $(\epsilon_{i}, \delta_{i})$-LDP guarantee for the $i$-th MT.
We can note that all MTs have their own privacy parameters $\epsilon_{i}$ and $\delta_{i}$.
We denote by $T$ the number of communication rounds, by $\boldsymbol{w}^{0}$ the initial global parameter, by $\sigma_{i}$ the STD of additive noises for the $i$-th MT, by $t$ the index of the current communication round and by $q$ the random sampling ratio ($q=K/U$).

At the beginning, the server broadcasts the initiate global parameter $\boldsymbol{w}^{0}$ and $T$ to all MTs.
Then, $K$ active MTs are chosen and train the model parameters by using local datasets with preset terminal conditions, respectively.
During the local training, the local gradients $\boldsymbol{g}^{t}_{i,m}, \forall m\in\{1,\ldots,\vert \mathcal{D}_{i}\vert\}$ are clipped by the threshold $C$.
After the local training, the $i$-th MT, $\forall i\in\mathcal{K}$, will add noises to the trained model parameters $\boldsymbol{w}^{t+1}_i$, in which $\sigma_{i}^{2}$ is the variance of artificial noises and this value is calculated by the $i$-th MT according to the privacy level $(\epsilon_{i}, \delta_{i})$, sampling ratio $q$ and the number of communication rounds $T$.
When all active MTs finish local training process, they are required to upload the noised parameters $\boldsymbol{\widetilde{w}}^{t+1}_i$ to the server for aggregation.

Then, the server updates the global parameters ${\boldsymbol{w}}^{t+1}$ by aggregating the local parameters integrated with different weights according to~\eqref{equ:aggregation} and broadcasts them to the MTs. The accuracy of each MT will be estimated based on the received global parameters $\boldsymbol{w}^{t+1}$ using local testing datasets.
If preset terminal conditions are not satisfied, all MTs will start the next round of training process based on these updated global parameters.
In detail, when the aggregation time reaches a preset number of communication rounds $T$, our UDP completes and returns $\boldsymbol{w}^{T}$.
Owing to the local perturbations, it will be difficult for the honest-but-curious server to infer private features of the $i$-MT.
In this case, in order to effectively protect the $i$-th MT privacy, the STD of additive noises will be analyzed according to the concept of $(\epsilon_{i},\delta_{i})$-LDP in the following sections.

\begin{algorithm}
\caption{User-level DP (UDP)}
\label{alg:UDP}
\LinesNumbered
\KwData{The number of communication rounds $T$, the initial global parameter $\boldsymbol{w}^{0}$, the sample ratio $q=K/U$, the clipping threshold $C$ and the LDP parameters $(\epsilon_{i}, \delta_{i})$ for all MTs}
{Initialize: $t = 0$}\\
{The server broadcasts $\boldsymbol{w}^{0}$ and $T$ to all MTs}\\
\While {$t < T$}
{
\For {$i\in \mathcal K$}
{
Update the local gradients:\\
$\boldsymbol{g}^{t}_{i,m}(\mathcal{D}_{i,m})=\nabla_{\boldsymbol{w}^{t}}{F_{i}(\mathcal{D}_{i,m}, \boldsymbol{w}^{t})}$, where $m\in\{1,\ldots,\vert \mathcal{D}_{i}\vert\}$\;
Clip the local gradients:\\
$\boldsymbol{g}^{t}_{i,m}(\mathcal{D}_{i,m}) = \boldsymbol{g}^{t}_{i,m}(\mathcal{D}_{i,m})/\max\left(1,\frac{\Vert\boldsymbol{g}^{t}_{i,m}(\mathcal{D}_{i,m})\Vert_{2}}{C}\right)$\;
Update the local parameters:
$\boldsymbol{w}^{t+1}_{i} = \boldsymbol{w}^{t}- \frac{1}{\vert\mathcal{D}_{i}\vert}\sum_{m=1}^{\vert\mathcal{D}_{i}\vert}\eta\boldsymbol{g}^{t}_{i,m}(\mathcal{D}_{i,m})$\;
Calculate $\sigma_{i}$ according to LDP parameters $(\epsilon_{i}, \delta_{i})$:\\
$\boldsymbol{\widetilde{w}}^{t+1}_{i}=\boldsymbol{w}^{t+1}_{i}+\mathcal N(0, \sigma_{i}^2\mathbf{I})$\;
upload noised parameters to the server\;
}
Update the global parameters $\boldsymbol{w}^{t+1}$ as\\
\quad\quad $\boldsymbol{w}^{t+1} = \sum\limits_{i\in \mathcal K}{p_{i}\widetilde{\boldsymbol{w}}^{t+1}_{i}}$\;
The server broadcasts the global parameters \\
\For {$\mathcal C_i\in \{\mathcal C_1, \mathcal C_2, \ldots,\mathcal C_{U}\}$}
{
Test the aggregating parameters $\boldsymbol{w}^{t+1}$\;
using local dataset
}
$t = t + 1$\;
}
\KwResult{$\boldsymbol{w}^{T}$}
\end{algorithm}

\subsection{Bound of the Moment}
Before calculating the STD $\sigma_{i}$ of the additive noises, we will first enhance the classic moments accountant method in~\cite{Abadi2016Deep}.
According to~\cite{Abadi2016Deep}, using Gaussian mechanism, we can define privacy loss of $i$-th MT after $T$ communication rounds in our UDP algorithm by
\begin{equation}
c \triangleq \exp\left(\alpha^{T}(\lambda)\right) = \exp\left(\sum_{t=1}^{T}\alpha(\lambda)\right),
\end{equation}
where the moment generating function $\alpha(\lambda)$ is given by
\begin{equation}\label{equ:moment_accountant}
\alpha(\lambda) \triangleq \ln(\max \{D_{\nu_{1},\nu_{0}}, D_{\nu_{0},\nu_{1}}\}),
\end{equation}
$\lambda$ is any positive integer, $\nu_{0}$ denotes the Gaussian probability density function (PDF) of $\mathcal{N}(0,\sigma_{i}^{2})$, $\nu_{1}$ denotes the mixture of two Gaussian distributions $q\mathcal{N}(\Delta \ell,\sigma_{i}^{2})+(1-q)\mathcal{N}(0,\sigma_{i}^{2})$, $q=K/U$ is the random sampling ratio in UDP algorithm (means that all MTs have the same probability to be selected in the aggregation) and $\Delta \ell$ denotes the sensitivity of the local training process $\ell$, respectively. The expression of $D_{\nu_{1},\nu_{0}}$ and $D_{\nu_{0},\nu_{1}}$ can be written as
\begin{equation}
D_{\nu_{1},\nu_{0}}=\mathbb{E}_{z\sim \nu_{1}}\left(\frac{\nu_{1}}{\nu_{0}}\right)^{\lambda}=\mathbb{E}_{z\sim \nu_{0}}\left(\frac{\nu_{1}}{\nu_{0}}\right)^{\lambda+1},
\end{equation}
and
\begin{equation}
D_{\nu_{0},\nu_{1}}=\mathbb{E}_{z\sim \nu_{0}}\left(\frac{\nu_{0}}{\nu_{1}}\right)^{\lambda}=\mathbb{E}_{z\sim \nu_{0}}\left(\frac{\nu_{1}}{\nu_{0}}\right)^{-\lambda}.
\end{equation}

However, the derivations in~\cite{Abadi2016Deep} for the bound of moments can only be applied for the rigorous constraint $q \leq \frac{1}{16\sigma_{i}}$.
To tackle this problem, we propose~\emph{\textbf{Lemma~\ref{lemma:Comp_div}}} to further bound the moment.
\begin{lemma}\label{lemma:Comp_div}
Considering two Gaussian distributions $\nu_{0}$ and $\nu_{1}$ used in the moments accountant method, they satisfy the following relationship:
\begin{equation}
D_{\nu_{1},\nu_{0}}\geq D_{\nu_{0},\nu_{1}}.
\end{equation}
\end{lemma}
\begin{IEEEproof}
See Appendix~\ref{appendix:Comp_div}.
\end{IEEEproof}

Note that the only difference between the $D_{\nu_{1},\nu_{0}}$ and $D_{\nu_{0},\nu_{1}}$ is the factor of $\lambda+1$ and $-\lambda$ on the exponent.
With~\emph{\textbf{Lemma~\ref{lemma:Comp_div}}}, we can directly obtain that $\alpha(\lambda) = D_{\nu_{1},\nu_{0}}$.
In this way, we can calculate the privacy loss by bound $D_{\nu_{1},\nu_{0}}$, which can relax the constraint $q \leq \frac{1}{16\sigma_{i}}$. In the following subsection, we will derive the STD of the Gaussian noises added in the UDP algorithm with~\emph{\textbf{Lemma~\ref{lemma:Comp_div}}}.
\subsection{User-level Noise Calculation}
With the sensitivity and the bound of the moment, we can design the Gaussian mechanism $\mathcal{N}(0, \sigma_{i})$ for the $i$-th MT with $(\epsilon_{i}, \delta_{i})$-LDP requirement in terms of the sampling ratio $q = K/U$ and the number of communication rounds $T$.
The STD of the Gaussian noises can be derived according to the following theorem.
\begin{theorem}\label{theorem:DP_ConvforK}
Given the sampling ratio $q$ and the number of communication rounds $T$, to guarantee $(\epsilon_{i}, \delta_{i})$-LDP for the $i$-th MT, the STD of noises $\sigma_{i}$ from Gaussian mechanism should satisfy
\begin{equation}\label{equ:DP_ConvforK}
\sigma_{i} =\frac{\Delta\ell\sqrt{2qT\ln(1/\delta_{i})}}{\epsilon_{i}},
\end{equation}
where $\Delta\ell$ is the sensitivity of the local training process and $\ell(\cdot)$ denotes the local training process.
\end{theorem}
\begin{IEEEproof}
In this proof, we first need to calculate the sensitivity of the local training process defined by~\eqref{eq:org_sensi}.
According to the definition of LDP, we consider two adjacent datasets $\mathcal{D}_{i}, \mathcal{D}'_{i}\in \mathcal{X}$ in the $i$-th MT, where $\mathcal{D}_{i}$ and $\mathcal{D}'_{i}$ have the same size, and only differ by one sample.
Consequently, for the $i$-th MT with the training dataset $\mathcal{D}_{i}$, the $t$-th local training process can be written into the following form:
\begin{equation}
\boldsymbol{w}_{i}^{t+1} = \ell(\mathcal{D}_{i}, \boldsymbol{w}^{t}),
\end{equation}
Therefore, the sensitivity of the local training process can be given as
\begin{equation}\label{equ:sensitivity_train}
\Delta \ell = \max\limits_{\mathcal{D}_{i},\mathcal{D}'_{i}\in \mathcal{X}}\Vert \ell(\mathcal{D}_{i}, \boldsymbol{w}^{t})-\ell(\mathcal{D}'_{i}, \boldsymbol{w}^{t})\Vert.
\end{equation}
Assuming that the batch size in the local training is equal to the number of training samples, we have
\begin{equation}\label{equ:sensitivity}
\begin{aligned}
\Delta \ell&=\frac{\eta}{\vert \mathcal{D}_{i}\vert}\sum_{m=1}^{\vert \mathcal{D}_{i}\vert}\max\limits_{\mathcal{D}_{i,m},\mathcal{D}'_{i,m}\in \mathcal{X}}\left\Vert\boldsymbol{g}^{t}_{i,m}(\mathcal{D}_{i,m})\right.\\
&\quad\left.-\boldsymbol{g}^{t}_{i,m}(\mathcal{D}'_{i,m})\right\Vert\leq\frac{2\eta C}{\vert \mathcal{D}_{i}\vert},
\end{aligned}
\end{equation}
where $C$ is the clipping threshold to bound $\Vert \boldsymbol{g}^{t}_{i,m}\Vert$.

Then, we need to calculate the privacy loss using the moments account method.
Hence, the $\lambda$-th moment $\alpha(\lambda_{n})$ can be expressed as~\eqref{equ:moment_accountant}.
Here, we want to bound $D_{\nu_{0}, \nu_{1}}$ and $D_{\nu_{1},\nu_{0}}$.
Based on~\textbf{Lemma~\ref{lemma:Comp_div}}, we only need to bound $D_{\nu_{1},\nu_{0}}$ and have
\begin{equation}
\begin{aligned}
&D_{\nu_{1},\nu_{0}}=\mathbb{E}_{z\sim \nu_{1}}\left(\frac{\nu_{1}}{\nu_{0}}\right)^{\lambda}\\
&=\int_{-\infty}^{+\infty} \nu_0\left(1-q+qe^{\frac{2z\Delta \ell-\Delta \ell^{2}}{2\sigma_{i}^{2}}}\right)^{\lambda+1} \mathrm{d}z\\
&=\int_{-\infty}^{+\infty} \nu_0
\sum_{l=0}^{\lambda+1}\left(\begin{matrix}\lambda+1\\ l\end{matrix}\right)(1-q)^{\lambda+1-l}q^le^{\frac{l\left(2z\Delta\ell-\Delta \ell^{2}\right)}{2\sigma_{i}^{2}}}\mathrm{d}z\\
&=
\sum_{l=0}^{\lambda+1}\left(\begin{matrix}\lambda+1\\ l\end{matrix}\right)(1-q)^{\lambda+1-l}q^{l}e^{\frac{l(l-1)\Delta \ell^{2}}{2\sigma_{i}^{2}}}\\
&\leq\left(1-q+qe^{\frac{\lambda\Delta \ell^{2}}{2\sigma_{i}^{2}}}\right)^{\lambda+1}\leq e^{q(\lambda+1)\left(e^{\frac{\lambda\Delta \ell^{2}}{2\sigma_{i}^{2}}}-1\right)}.\\
\end{aligned}
\end{equation}
Assuming $\frac{\lambda\Delta\ell^{2}}{2\sigma_{i}^{2}}\ll1$ for $\lambda \in [1,T]$, we have
\begin{equation}\label{equ:divergence_bound}
D_{\nu_{1},\nu_{0}} \leq e^{q(\lambda+1)\left(\frac{\lambda\Delta\ell^{2}}{2\sigma_{i}^{2}}+O\left(\frac{\lambda^{2}\Delta \ell^{4}}{4\sigma_{i}^{4}}\right)\right)}\approx e^{\frac{q\lambda(\lambda+1)\Delta\ell^{2}}{2\sigma_{i}^{2}}}.
\end{equation}
We use the above moments and have
\begin{equation}\label{equ:moments}
\alpha^{T}(\lambda) \leq \sum_{t=1}^{T}\alpha(\lambda, \sigma_{i})=\frac{Tq\lambda(\lambda+1)\Delta\ell^{2}}{2\sigma_{i}^{2}}.
\end{equation}
Using the tail bound by moments~\cite{Abadi2016Deep}, we have
\begin{equation}\label{appendix:A_1}
\begin{aligned}
\delta_{i}=\min_{\lambda}\exp\left(\frac{Tq\lambda(\lambda+1)\Delta\ell^2}{2\sigma_{i}^2}-\lambda\epsilon_{i}\right).
\end{aligned}
\end{equation}
Considering inequation~\eqref{appendix:A_1}, we know
\begin{multline}
\frac{Tq\lambda(\lambda+1)\Delta\ell^2}{2\sigma_{i}^2}-\lambda\epsilon_{i}=\frac{Tq\Delta\ell^{2}}{2\sigma_{i}^2}\left(\lambda+\frac{1}{2}-\frac{\epsilon_{i}\sigma_{i}^2}{Tq\Delta\ell^2}\right)^2\\
-\frac{Tq\Delta s^{2}}{2\sigma_{i}^2}\left(\frac{1}{2}-\frac{\epsilon_{i}\sigma_{i}^2}{Tq\Delta\ell^2}\right)^2.
\end{multline}
When setting $\lambda = \frac{\epsilon_{i}\sigma_{i}^2}{Tq\Delta\ell^2}-\frac{1}{2}$, we have
\begin{equation}
\begin{aligned}
\ln\left(\frac{1}{\delta_{i}}\right)&\leq\frac{Tq\Delta\ell^{2}}{2\sigma_{i}^2}\left(\frac{1}{2}-\frac{\epsilon_{i}\sigma_{i}^2}{Tq\Delta\ell^2}\right)^2\\
&=\frac{Tq\Delta\ell^{2}}{8\sigma_{i}^2}-\frac{\epsilon_{i}}{2}+\frac{\epsilon_{i}^2 \sigma_{i}^2}{2Tq\Delta\ell^{2}}.
\end{aligned}
\end{equation}
Since $\delta_{i} \in (0,1)$, we can obtain
\begin{equation}\label{appendix:A_2}
\frac{Tq\lambda(\lambda+1)\Delta\ell^2}{2\sigma_{i}^2}-\lambda\epsilon_{i}<0.
\end{equation}
Combine inequation~\eqref{appendix:A_2}, we can bound $\ln\left(1/\delta\right)$ as
\begin{equation}
\ln\left(\frac{1}{\delta_{i}}\right)<-\frac{\epsilon_{i}}{4}+\frac{\epsilon_{i}^2 \sigma_{i}^2}{2Tq\Delta\ell^{2}}<\frac{\epsilon_{i}^2 \sigma_{i}^2}{2Tq\Delta \ell^{2}}.
\end{equation}
Therefore, we can choose $\sigma_{i}$ satisfies~\eqref{equ:DP_ConvforK} to guarantee $(\epsilon_{i}, \delta_{i})$-DP in the FL framework.
\end{IEEEproof}

\emph{\textbf{Theorem~\ref{theorem:DP_ConvforK}}} quantifies the relation between the noise level $\sigma_{i}$ and the privacy level $\epsilon_{i}$. It shows that for a fixed perturbation $\sigma_{i}$ on model parameters, a larger $q$ leads to a weaker privacy guarantee (i.e., a larger $\sqrt{q}/\epsilon_{i}$).
This is indeed true since when more MTs are involved in computing $\boldsymbol{w}$ at each communication round, there will be a larger probability of information leakage for each MT.
Also, for a given $\epsilon_{i}$, a larger $T$ in the total training process lead to a higher chance of information leakage because the observer may obtain more information for training datasets.
Furthermore, for a given privacy protection level $\epsilon_{i}$ and $T$, a larger value of $q$ leads to a larger value of $\sigma_{i}$, which helps reduce the $i$-th MT concerns on participating in FL because the $i$-th MT are allowed to add more noise to the trained local models.
This requires us to choose the parameters carefully in order to have a reasonable privacy level.
\section{Discounting Method in UDP}\label{Sec:dyn_alo}
In this section, we first introduce the property that there exists an optimal number of communication rounds in the UDP algorithm.
Based on this property, we propose a CRD method to improve convergence performance in the training process.
Then, we provide a noise calculation method for obtaining the STD of the additive noises in the case of varying $T$ during the FL training.
Finally, we will summarize our proposed CRD algorithm.

\subsection{Performance Analysis for UDP}

First, we start with the essential assumption of on the global loss function $F(\cdot)$ defined by $F(\cdot)\triangleq \sum_{i=1}^{U}p_iF_{i}(\cdot)$, and the $i$-th local loss function $F_{i}(\cdot)$ for the analysis, which can be satisfied normally.
\begin{assumption}\label{assu:conv_FL}
We assume the following conditions for the loss function of all MTs:
\begin{enumerate}
\item[\emph{1)}] $F_{i}(\boldsymbol{w})$ is convex;
\item[\emph{2)}] $F_{i}(\boldsymbol{w})$ satisfies the Polyak-Lojasiewicz condition with the positive parameter $\mu$, which implies that
$F(\boldsymbol{w})-F(\boldsymbol{w}^{*})\leq\frac{1}{2\mu}\Vert \nabla F(\boldsymbol{w})\Vert^{2}$, where $\boldsymbol{w}^{*}$ is the optimal result;
\item[\emph{3)}] $F_{i}(\boldsymbol{w})$ is $L$-Lipschitz smooth, i.e., for any $\boldsymbol{w}$, $\boldsymbol{w}'$, $\Vert \nabla F_{i}(\boldsymbol{w})-\nabla F_{i}(\boldsymbol{w}')\Vert\leq L\Vert \boldsymbol{w}-\boldsymbol{w}'\Vert$, where $L$ is a constant determined by the practical loss function;
\item[\emph{4)}] $\eta \leq \frac{1}{L}$, where $\eta$ is the learning rate;
\item[\emph{5)}] For any $i$ and $\boldsymbol{w}$, $\Vert\nabla F_{i}(\boldsymbol{w})-\nabla F(\boldsymbol{w})\Vert\leq \varepsilon_{i}$ and $\mathbb{E}\{\varepsilon_{i}\}=\varepsilon$, where $\varepsilon_{i}$ is the divergence metric.
\end{enumerate}
\end{assumption}

Based on the above assumptions and~\emph{\textbf{Theorem~1}}, we can obtain the following result which characterizes the convergence performance of the UDP algorithm.
\begin{theorem}\label{theorem:loss_ConvforK}
To guarantee $(\epsilon_{i}, \delta_{i})$-DP for all MTs, the convergence upper bound of the UDP algorithm after $T$ communication rounds is given by
\begin{multline}\label{equ:ConvforK_loss}
\mathbb{E}\{F(\boldsymbol{w}^{T})\}-F(\boldsymbol{w}^{*})\leq A^{T}(F(\boldsymbol{w}^{0})-F(\boldsymbol{w}^{*}))\\
+(1-A^T)\left(\frac{\kappa_{0}TK}{U^2}\sum_{i=1}^{U}\frac{\ln(1/\delta_{i})}{\epsilon_{i}^{2}}+\frac{\kappa_{1}U(U-K)}{K(U-1)}\right),
\end{multline}
where $A = 1-2\mu\eta+\mu\eta^{2}L$, $\kappa_{0} = \frac{L^2\Delta\ell^2}{\mu}$ and $\kappa_{1} = \frac{\eta^{2} L^{2}\varepsilon}{2\mu}$.
\end{theorem}
\begin{IEEEproof}
See Appendix ~\ref{appendix:ConvforK}.
\end{IEEEproof}

From \emph{\textbf{Theorem~\ref{theorem:loss_ConvforK}}}, we can find an explicit tradeoff between convergence performance and privacy: When privacy guarantee is weak (large values of $\epsilon_{i}$ and $\delta_{i}$, and small values of $\sum_{i=1}^{U}\frac{\ln(1/\delta_{i})}{\epsilon_{i}^{2}}$), the convergence bound will be small, which indicates a tight convergence to the optimal weights.
From our assumptions, we know that $\eta L\leq 1$ and obtain $A< 1$.
Then, we show the process on discovering the relationship between the convergence upper bound and the number of total MTs $U$, the number of participant MTs $K$ and the number of communication rounds $T$.
We find that the number of communication rounds $T$ is a key factor, and then obtain the following theorem.

\begin{theorem}\label{theorem:Covex_CR}
There exists an optimal number of communication rounds to achieve the best learning performance for the given $\epsilon_{i}$, $\forall i\in\mathcal{U}$, and a sufficiently large $U$.
\end{theorem}
\begin{IEEEproof}
With a slight abuse of notation, we consider continuous values of $U\geq 1$, $1\leq K \leq U$ and $T\geq 1$.
Let $h(U, K, T)$ denote the right hand side (RHS) of~\eqref{equ:ConvforK_loss} and we have
\begin{multline}\label{Par_1}
\frac{\partial^{2} h(U, K, T,\epsilon_{i})}{\partial T^{2}}=A^{T}\ln^{2}{A}\bigg{(}F(\boldsymbol{w}^{0})-F(\boldsymbol{w}^{*})\\
-\frac{\kappa_{0}TK}{U^{2}}\sum_{i=1}^{U}\frac{\ln(1/\delta_{i})}{\epsilon_{i}^{2}}-\frac{\kappa_{1}(U-K)}{K(U-1)}-\frac{\kappa_{0}TK}{U\ln{A}}\sum_{i=1}^{U}\frac{\ln(1/\delta_{i})}{\epsilon_{i}^{2}}\bigg{)}.
\end{multline}

It can be seen that the first term and fourth term of on the RHS of~\eqref{Par_1} are always positive.
When $U$ and $K$ are set to be large enough, and $\sum_{i=1}^{U}\frac{\ln(1/\delta_{i})}{\epsilon_{i}^{2}}$ is small (proper privacy guarantee), we can see that the second term of on the RHS of~\eqref{Par_1} is small.
In this case, we have $\frac{\partial^{2} h(U, T,\epsilon_{i})}{\partial T^{2}}>0$ and the upper bound is convex for $T$.

Then we consider the condition $K = U$, we have
\begin{multline}\label{Par_3}
\frac{\partial^{2} h(U, T, \epsilon_{i})}{\partial T^{2}}=A^{T}\ln^{2}{A}\bigg{(}F(\boldsymbol{w}^{0})-F(\boldsymbol{w}^{*})\\
-\frac{\kappa_{0}T}{U}\sum_{i=1}^{U}\frac{\ln(1/\delta_{i})}{\epsilon_{i}^{2}}-\frac{\kappa_{0}T}{U\ln{A}}\sum_{i=1}^{U}\frac{\ln(1/\delta_{i})}{\epsilon_{i}^{2}}\bigg{)}.
\end{multline}
If $U$ is set to be large enough, and $\sum_{i=1}^{U}\frac{\ln(1/\delta_{i})}{\epsilon_{i}^{2}}$ is small, we have $\frac{\partial^{2} h(U, T,\epsilon_{i})}{\partial T^{2}}>0$ and the upper bound is convex for $T$.
\end{IEEEproof}

In more detail, a larger $T$ has a negative impact on the model quality by increasing the amount of noise added in each communication round for a given $\epsilon_{i}$, $\forall i\in\mathcal{U}$ (in line with \eqref{equ:ConvforK_loss}), but it also has a positive impact on the convergence because it reduces the loss function value with more iterations.

According to our analysis above, there exists an optimal value of $T$ for given privacy levels $\epsilon_{i}$, $\forall i\in\mathcal{U}$.
However, this optimal value cannot be derived directly, since some parameters of the loss function are difficult to obtain accurately.
One possible method for obtaining the optimal $T$ is through exhaustive search, i.e., try different value of $T$ and choose the one with the highest convergence performance as the practical value in use.
The time complexity of the exhaustive search method is determined by the searching interval and a smaller interval means a larger complexity but a higher performance.
Hence, exhaustive search is time-consuming and computationally complex.
In the next section, we will propose an efficient algorithm for finding a good value of $T$ to achieve a high convergence performance.
\subsection{Proposed Discounting Communication Rounds}
\begin{figure*}[ht]
\centering
\includegraphics[width=6.5in,angle=0]{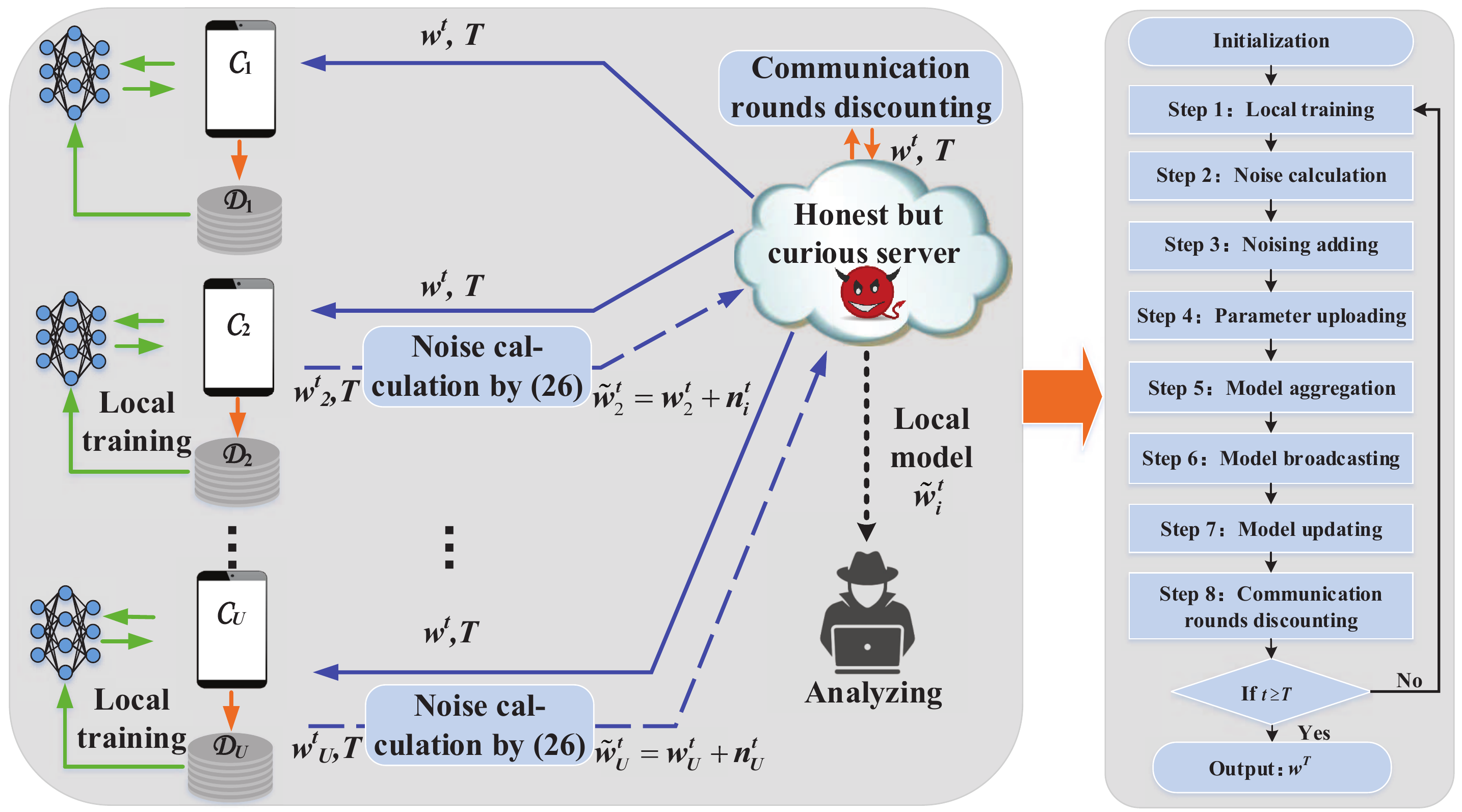}
\caption{The training process of our proposed UDP algorithm with CRD method at the $t$-th communication round. As shown in the figure, we have $8$ steps in each communication round. Our algorithm will recalculate the value of $T$ (the updated $T$ is generally less than the previous one) when the training performance stops improving. The advantages of our algorithm are as follows. First, we can obtain a smaller $T$, i.e., less training time, compared with the conventional UDP-based FL. Second, we can achieve a better convergence performance (due to less noise added to the model) while keeping the DP level unchanged.}
\label{fig:CRD_method}
\end{figure*}
Based on the analysis above, we can note that an improper value of $T$ will damage the performance of the UDP algorithm.
Hence, if we reduce the value of $T$ slightly when the training performance stops improving, we can obtain a smaller STD as well as improving the training performance.
Therefore, we design a CRD algorithm by adjusting the number of communication rounds $T$ with a discounting method during the training process to achieve a better convergence performance.
The training process of such a CRD algorithm in UDP contains following steps:
\begin{itemize}
\item[$\bullet$] \textbf{Initialization}: The server broadcasts the initial parameters (i.e., $\boldsymbol{w}^{0}$ and $T$) to all MTs;
\item[$\bullet$] \textbf{Step 1}: \emph{Local training: } All active MTs locally compute training parameters with local datasets and the global parameter. In order to prove the DP guarantee, the influence of each individual example on local gradients should be bounded with a clipping threshold $C$. Each gradient vector will be clipped in $L_{2}$ norm, i.e., the $i$-th local gradient vector $\boldsymbol{g}_{i}^{t}$ at the $t$-th communication round is replaced by $\boldsymbol{g}_{i}^{t}/\max(1, \Vert \frac{\boldsymbol{g}_{i}^{t}}{C}\Vert)$. We can remark that parameter clipping of this form is a popular ingredient of ML for non-privacy reasons;
\item[$\bullet$] \textbf{Step 2}: \emph{Noise calculation: } Each MT obtain the STD $\sigma_{i}$ of artificial Gaussian noise using the proposed noise calculation method (introduced in the following subsection);
\item[$\bullet$] \textbf{Step 3}: \emph{Noise adding: } Each MT add artificial Gaussian noise with a certain STD to the local trained parameters in order to guarantee $(\epsilon_{i}, \delta_{i})$-LDP;
\item[$\bullet$] \textbf{Step 4}: \emph{Parameter uploading: } All active MTs upload the noised parameters to the server for aggregation;
\item[$\bullet$] \textbf{Step 5}: \emph{Model aggregation:} The server performs aggregation over the uploaded parameters from MTs;
\item[$\bullet$] \textbf{Step 6}: \emph{Model broadcasting: }The server broadcasts the aggregated parameters and the number of communication rounds $T$ to all MTs;
\item[$\bullet$] \textbf{Step 7}: \emph{Model updating: }All MTs update their respective models with the aggregated parameters, then test the performance of the updated models and upload the performance to the server;
\item[$\bullet$] \textbf{Step 8}: \emph{Communication rounds discounting: } When the convergence performance stops improving by the following decision $ \mathcal{V}(\boldsymbol{w}^{t})-\mathcal{V}(\boldsymbol{w}^{t+1})<\zeta$ and $\zeta$ is the threshold, the discounting method will be triggered in the server, where $ \mathcal{V}(\boldsymbol{w}^{t})$ is the test loss by the model $\boldsymbol{w}^{t}$. The server will obtain a smaller $T$ than the previous one with a linear discounting factor $\beta$ and an integer value by $T=\lfloor\beta (T-t)\rfloor+t$. This factor can control the decaying speed of $T$. The FL process will be completed when the aggregation time reaches the preset $T$.
\end{itemize}

First, we can note that the value of $\zeta$ will have an impact on the number of communications $T$ and not affect the convergence and the privacy guarantee directly.
Then, the value of $\zeta$ is to predict whether the training process is stopping and determine when to trigger the discounting method (adjust the number of communications).
In the conventional machine learning, a threshold is common to predict whether the training process is stopping.
Therefore, in our experiments, we adopt a small positive value $0.001$ as the value of $\zeta$.
Moreover, in this method, the value of $T$ is determined iteratively to ensure a high convergence performance in FL training.
Obviously, when the value of $T$ is adjusted, we must calculate a new STD of additive noises in terms of previous training process.
The diagrammatic expression of this method is shown in Fig.~\ref{fig:CRD_method}.
Therefore, we will develop a noise calculation method to update the STD of additive noises and $T$ alternately in the following subsection.

\subsection{Noise Calculation for Varying $T$}\label{Sec:DP_compos}
Now, let $t$ be the index of the current communication round and $\sigma^{\tau}_{i}$ ($0\leq \tau \leq t-1$) be the STD of additive noises for the $i$-th MT at the $\tau$-th communication round.
In our CRD algorithm, with a new $T$, if $t$ is greater than $T$, the training process will stop.
If $t$ is less than $T$, we need to calculate the STD of noises and add them on the local parameters in the following communication round.
Considering this, we obtain the following theorem.
\begin{theorem}
\label{theorem:recal_noise}
After $t$ ($0\leq t < T$) communication rounds and a new $T$, the STD of additive noises for the $i$-th MT to guarantee $(\epsilon_{i}, \delta_{i})$-LDP can be given as
\begin{equation}\label{equ:recal_noise}
\sigma^{t}_{i} = \left(\frac{T-t}{\frac{\epsilon_{i}^2}{2q\Delta \ell^2\ln\left(\frac{1}{\delta_{i}}\right)}-\sum_{\tau=0}^{t-1}\frac{1}{(\sigma^{\tau}_{i})^{2}}}\right)^{\frac{1}{2}}.
\end{equation}
\end{theorem}
\begin{IEEEproof}
Using the above definition of moments~\eqref{equ:moments} and a preset value of $T$, we have
\begin{multline}
\alpha^{T}(\lambda) \leq \sum_{\tau=1}^{T}\alpha(\lambda)=\sum_{\tau=1}^{t}\frac{q\lambda(\lambda+1)\Delta \ell^{2}}{2(\sigma_{i}^{\tau})^{2}}\\
+\sum_{\tau=1}^{T-t}\frac{q\lambda(\lambda+1)\Delta \ell^{2}}{2\sigma_{i}^{2}}\\
=\sum_{\tau=1}^{t}\frac{q\lambda(\lambda+1)\Delta \ell^{2}}{2(\sigma_{i}^{\tau})^{2}}+\frac{(T-t)q\lambda(\lambda+1)\Delta \ell^{2}}{2\sigma_{i}^{2}}.
\end{multline}
Using the tail bound by moments~\cite{Abadi2016Deep}, we have
\begin{equation}
\delta = \min_{\lambda}\exp\left(\alpha^{T}(\lambda)-\lambda\epsilon_{i}\right).
\end{equation}
Because of
\begin{multline}
\min\{\alpha^{T}(\lambda)-\lambda\epsilon_{i}\} =-\frac{q\Delta \ell^2}{8}\left(\sum_{\tau=1}^{t}\frac{1}{(\sigma_{i}^{\tau})^{2}}+\frac{T-t}{\sigma_{i}^{2}}\right)\\
+\frac{\epsilon_{i}}{2}-\frac{\epsilon_{i}^2}{2q\Delta \ell^2\left(\sum_{\tau=1}^{t}\frac{1}{(\sigma_{i}^{\tau})^{2}}+\frac{T-t}{\sigma_{i}^{2}}\right)},
\end{multline}
where $\lambda = -\frac{1}{2}+\frac{\epsilon_{i}\sigma_{i}^2}{q\Delta \ell^2\left(\sum_{\tau=1}^{t}\frac{\sigma_{i}^2}{(\sigma_{i}^{\tau})^{2}}+T-t\right)}$.
Therefore, we have
\begin{equation}\label{equ:bound_delta}
\begin{aligned}
\ln\left(\frac{1}{\delta_{i}}\right)&\leq\frac{q\Delta \ell^2}{8}\left(\sum_{\tau=1}^{t}\frac{1}{(\sigma_{i}^{\tau})^{2}}+\frac{T-t}{\sigma_{i}^{2}}\right)-\frac{\epsilon_{i}}{2}\\
&\quad+\frac{\epsilon_{i}^2}{2q\Delta \ell^2\left(\sum_{\tau=1}^{t}\frac{1}{(\sigma_{i}^{\tau})^{2}}+\frac{T-t}{\sigma_{i}^{2}}\right)}\\
&<\frac{\epsilon_{i}^2}{2q\Delta \ell^2\left(\sum_{\tau=1}^{t}\frac{1}{(\sigma_{i}^{\tau})^{2}}+\frac{T-t}{\sigma_{i}^{2}}\right)}.
\end{aligned}
\end{equation}
Then, we can set~\eqref{equ:recal_noise} to guarantee $(\epsilon_{i}, \delta_{i})$-DP for the following training.

\end{IEEEproof}

In~\emph{\textbf{Theorem~\ref{theorem:recal_noise}}}, we can obtain a proper STD of the additive noises based on the previous training process and the value of $T$.
From this result, we can find that if we have large STDs (strong privacy guarantee) in the previous $t-1$ training processes, i.e., $\sigma^{\tau}_{i}$ is large, the calculated STD will be small (weak privacy guarantee), i.e., $\sigma^{t}_{i}$ is small.

We can also note that if the value of $T$ is not changed in this communication round, the value of STD will remain unchanged.
Considering $\sigma^{t}_{i}$, $\sigma^{t+1}_{i}$ and unchanged $T$, from equation~\eqref{equ:recal_noise}, we can obtain
\begin{equation}\label{equ:recal_noise_0}
\sigma^{t+1}_{i} = \left(\frac{T-t-1}{\frac{\epsilon_{i}^2}{2q\Delta \ell^2\ln\left(\frac{1}{\delta_{i}}\right)}-\sum_{\tau=0}^{t}\frac{1}{(\sigma^{\tau}_{i})^{2}}}\right)^{\frac{1}{2}},
\end{equation}
and
\begin{equation}\label{equ:recal_noise_1}
\frac{\epsilon^2_{i}}{2q\Delta\ell^2\ln\left(\frac{1}{\delta_{i}}\right)}-\sum_{\tau=0}^{t-1}\frac{1}{(\sigma^{\tau}_{i})^{2}}=\frac{T-t}{\left(\sigma^{t}_{i}\right)^{2}}.
\end{equation}
Substituting equation~\eqref{equ:recal_noise_1} into equation~\eqref{equ:recal_noise_0}, we have $\sigma^{t+1}_{i} = \sigma^{t}_{i}$, which is in line with our analysis.
In this case, we summarize the detailed steps of the UDP with CRD method in~\textbf{Algorithm~2}.
\begin{algorithm}
  \label{alg:CRD}
  \caption{UDP with CRD Method}
  \KwIn{The value of an initial $T$, LDP parameters $(\epsilon_{i},\delta_{i})$, clipping threshold $C$ and discounting factor $\beta$ ($\beta<1$).}
  {Initialize: $t = 0$ and $\boldsymbol{w}^{0}$}\\
  \While{$t < T$}
  {
    {Broadcast: $\boldsymbol{w}^{t}$ and $T$ to all MTs}\\
    \For{$\forall i\in \mathcal{K}$}
    {
    Calculate the STD of additive noises using~\eqref{equ:recal_noise}\;
    Locally train with clipping gradients:\\
    $\boldsymbol{w}_{i}^{t+1}= \ell(\mathcal{D}_{i}, \boldsymbol{w}^{t})$\;
    Add $(\epsilon_{i}, \delta_{i})$-LDP noise:\\
    $\boldsymbol{\widetilde{w}}_{i}^{t+1}= \boldsymbol{w}_{i}^{t}+\mathcal{N}(0, \sigma^{t}_{i}\textbf{I})$\;
    Upload noised parameters to the server\;
    }
    Aggregate received model parameters:\\
    $\boldsymbol{w}^{t+1}= \sum\limits_{i\in \mathcal{K}}p_{i}\boldsymbol{\widetilde{w}}_{i}^{t+1}$\;
    \For {$\mathcal C_i\in \{\mathcal C_1, \mathcal C_2, \ldots,\mathcal C_{U}\}$}{
    Test the aggregating parameters $\boldsymbol{w}^{t+1}$\;
    using local dataset}
    \If{$ \mathcal{V}(\boldsymbol{w}^{t})-\mathcal{V}(\boldsymbol{w}^{t+1})<\zeta$}
    {
    Update the preset value of $T$:\\
    $T=\lfloor\beta (T-t)\rfloor+t$\;
    }
    $t= t+1$\;
  }
  return $\boldsymbol{w}^{T}$
\end{algorithm}
\subsection{Complexity Analysis}

The main difference between the proposed training protocol in~\textbf{Algorithm 1} and the conventional algorithm is the gradients clipping, noise computing and noise adding.
During the training process in~\textbf{Algorithm 1}, each MT needs to clip the gradients by the equation $\boldsymbol{g}^{t}_{i,m}(\mathcal{D}_{i,m}) = \boldsymbol{g}^{t}_{i,m}(\mathcal{D}_{i,m})/\max\left(1,\frac{\Vert\boldsymbol{g}^{t}_{i,m}(\mathcal{D}_{i,m})\Vert_{2}}{C}\right)$ locally, which takes $O(s({\boldsymbol{w}}))$ time, where $s({\boldsymbol{w}})$ is the size of $\boldsymbol{w}$.
We can note that this clipping process will compare the norm of gradients and the clipping threshold $C$ to bound the gradients.
The noise computing can be expressed as $\sigma_{i} =\frac{\Delta\ell\sqrt{2qT\ln(1/\delta_{i})}}{\epsilon_{i}}$ and takes $O(1)$ time, which can be obtained directly when all required parameters are ready.
Finally, the noise adding is given by $\boldsymbol{\widetilde{w}}^{t+1}_{i}=\boldsymbol{w}^{t+1}_{i}+\mathcal N(0, \sigma_{i}^2\mathbf{I})$, where the complexity is governed by the size of model parameters and is $O(s({\boldsymbol{w}}))$.

Meanwhile, \textbf{Algorithm 2} has improved the noise computing and added the adjusting process for the number of communication rounds ($T$) compared with \textbf{Algorithm 1}.
Therefore, the noise computing in \textbf{Algorithm 2} can be given by $\sigma^{t}_{i}= (T-t)^{\frac{1}{2}}/\left(\frac{\epsilon^2_{i}}{2q\Delta \ell^2\ln(\frac{1}{\delta_{i}})}-\sum\limits_{\tau=0}^{t-1}\frac{1}{(\sigma^{\tau}_{i})^{2}}\right)^{\frac{1}{2}}$ and obtained directly when all required parameters are ready.
Moreover, the adjusting process for the number of communication rounds $T$ only needs few computing resources.
Both processes only need $O(1)$ time.

In summary, the time complexity analysis of the proposed training protocols, i.e., \textbf{Algorithm 1} and \textbf{Algorithm 2}, is $O(s({\boldsymbol{w}}))$. Furthermore, the proposed training protocols consume few more computing resources.
\section{Experimental Results}\label{Sec:Exm_Res}
In this section, we evaluate the accuracy of our analytical results for different learning tasks.
Then, we evaluate our proposed CRD method in UDP algorithm, and demonstrate the effectiveness of various parameter settings, such as the privacy level, the initial value of $T$ and discounting factor.
\subsection{Evaluation of Numerical Results}\label{subsec:analytical_results}
\begin{figure}[ht]
\centering
\includegraphics[width=3.0in,angle=0]{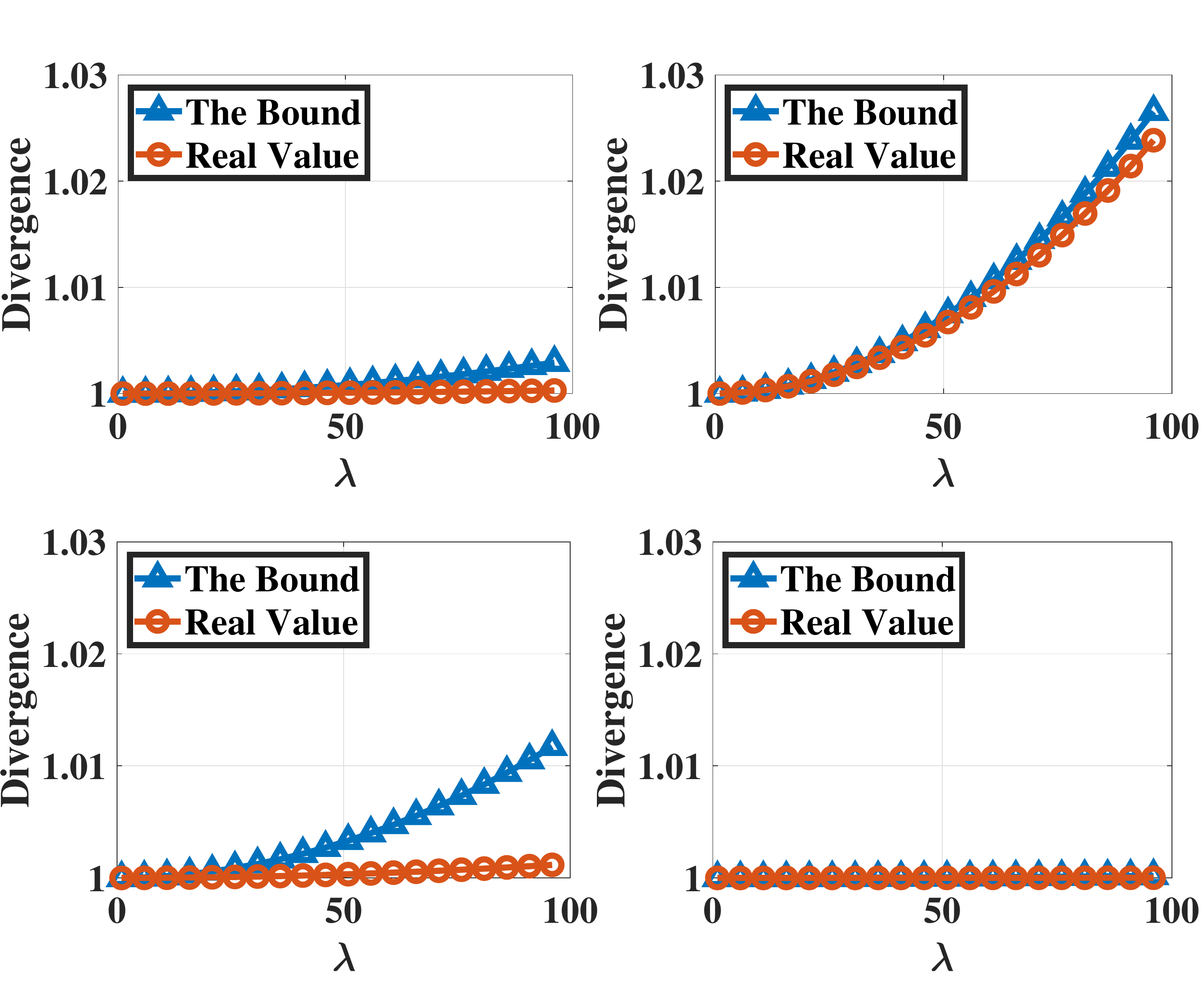}
\caption{The comparison of our bound and real value of divergence in~\eqref{equ:divergence_bound}. (a) $q=0.9$, $\vert \mathcal{D}_{i}\vert=800$, $U=50$. (b) $q=0.1$, $\vert \mathcal{D}_{i}\vert=800$, $U=50$. (c) $q=0.9$, $\vert \mathcal{D}_{i}\vert=400$, $U=50$. (d) $q=0.9$, $\vert \mathcal{D}_{i}\vert=800$, $U=200$.}
\label{fig:Comp_bound}
\end{figure}
In this subsection, we first describe our numerical validation of the bound compared with the real value in~\eqref{equ:divergence_bound} with $\sigma_{i} = 0.01$ by varying $\lambda$ from $1$ to $100$.
In Fig.~\ref{fig:Comp_bound} (a), we set  $q=0.9$, $\vert \mathcal{D}_{i}\vert=800$, $\forall i$, and $U=50$.
In our validation, our bound is close but always higher than the real divergence.
We also did tests for cases such as a smaller sampling ratio $q=0.1$, a smaller size of local datasets $\vert \mathcal{D}_{i}\vert=400$, $\forall i$, and a larger number of MTs $U=200$ in Fig.~\ref{fig:Comp_bound} (b), (c) and (d), respectively.
We find that the empirical bound always holds and closes under the given conditions, especially for a large enough $\vert \mathcal{D}_{i}\vert$.
Therefore, we have the conjecture that our bound is a valid for the analytical moments accountant of sampled Gaussian mechanism and seek its formal proof in our future work.

\subsection{Experimentation Setup}
\begin{figure*}[ht]
\centering
\includegraphics[width=5.5in,angle=0]{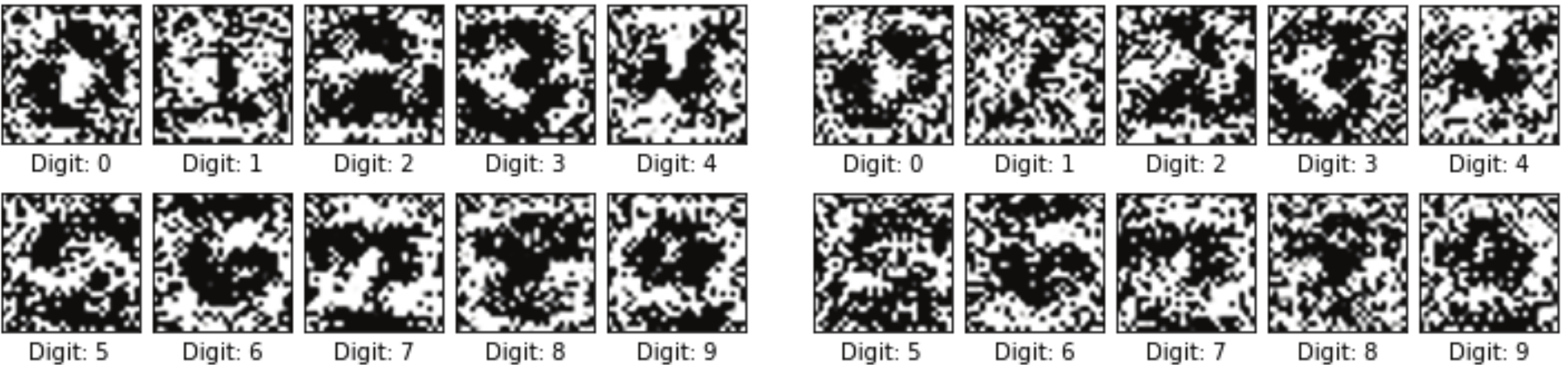}
\caption{Visual illustration of the standard MNIST dataset via MLP, in which the left is original FL model and the right is the model trained by UDP algorithm. Basically speaking, these subfigures show the typical digits that are learned by the machine models with the original FL generates less noisy digits than the UDP based FL.}
\label{fig:digits_visual_mlp}
\end{figure*}

\begin{figure}[ht]
\centering
\includegraphics[width=2.85in,angle=0]{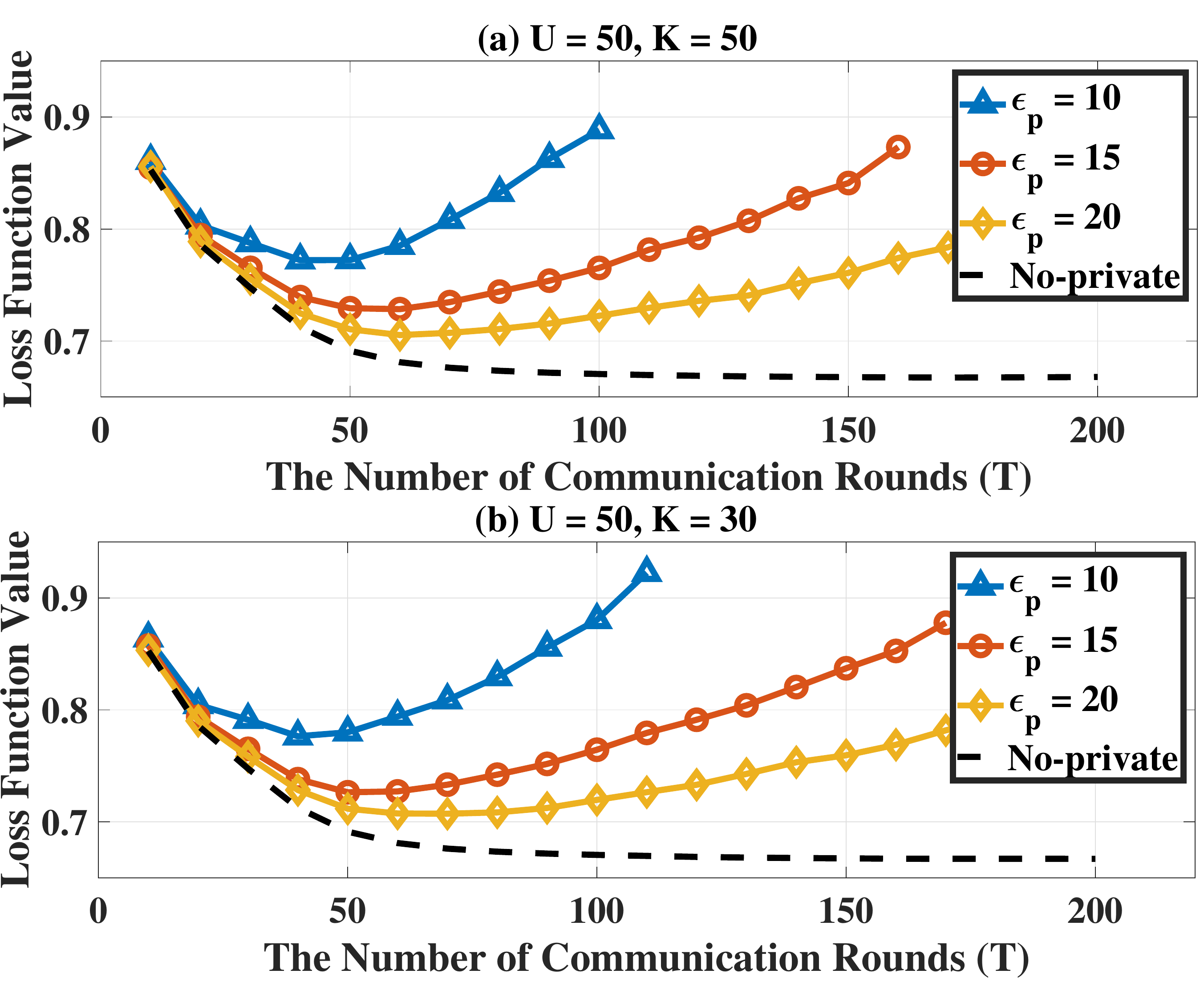}
\caption{Value of the loss function under various $T$ using the UDP algorithm with the SVM model. (a) $U=50, K=50$ ($q=1$). (b) $U=50, K=30$ ($q=0.6$).}
\label{fig:SVM_ConvforNandK_T}
\end{figure}
\begin{figure}[ht]
\centering
\includegraphics[width=2.85in,angle=0]{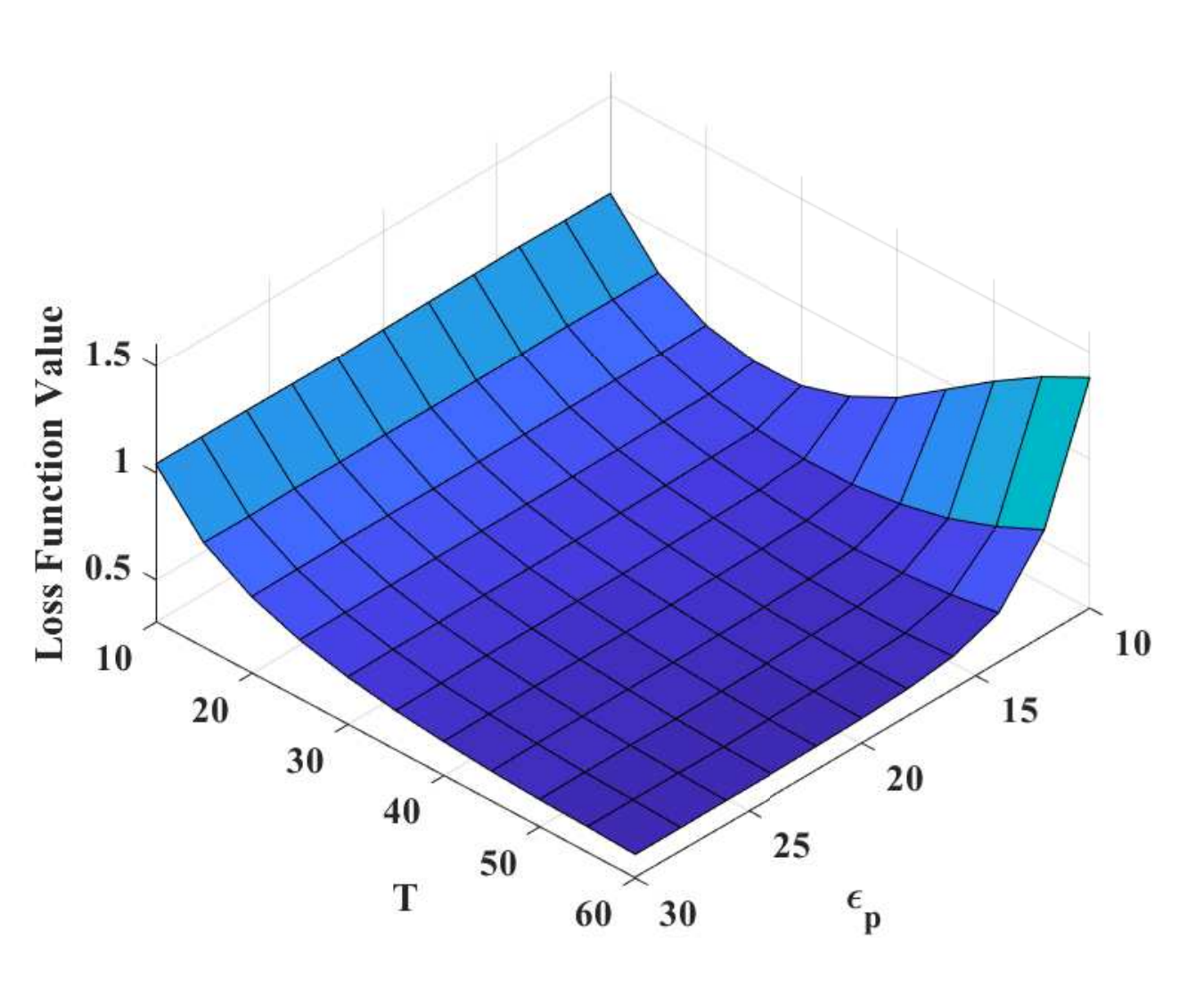}
\caption{Value of the loss function under various $T$ and privacy levels using the NN model based UDP algorithm with $U=K=50$ ($q=1$).}
\label{fig:ConvforN_T_eps_td}
\end{figure}
\begin{figure}[ht]
\centering
\includegraphics[width=2.85in,angle=0]{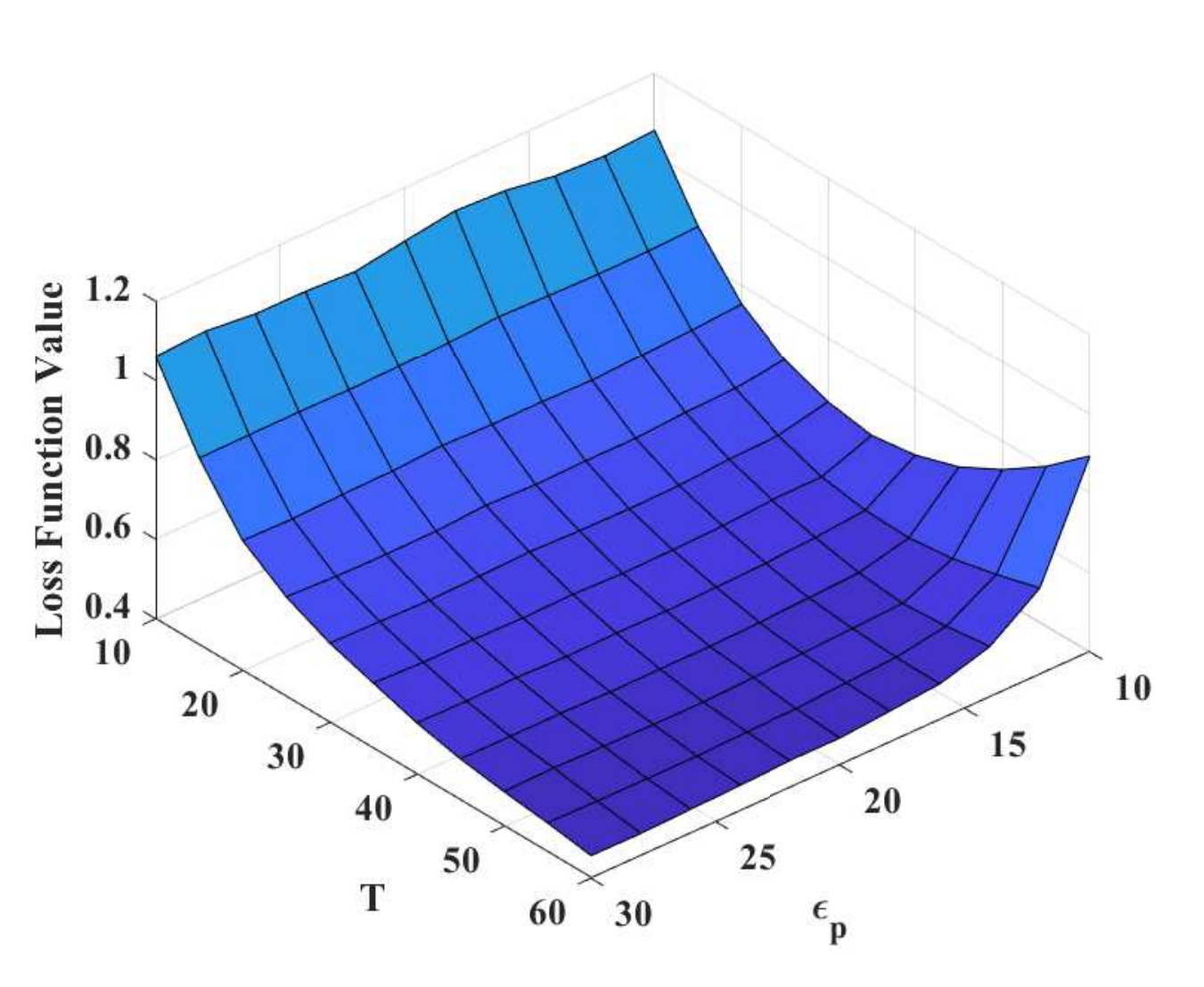}
\caption{Value of the loss function under various $T$ and privacy levels using the NN model based UDP algorithm with $U=50$ and $K=30$ ($q=0.6$).}
\label{fig:ConvforK_T_eps_td}
\end{figure}

We evaluate the training of three different machine learning models on different datasets, namely support vector machine (SVM) on the IPUMS-US dataset, multi-layer perceptron (MLP) on the standard MNIST dataset and the convolutional neural network (CNN) on the CIFAR-10 dataset, respectively.

\textbf{Models and Datasets Description.} The models include SVM, MLP and CNN, which are detailed as follows.

\emph{1)} SVM is trained on the IPUMS-US dataset, which are census data extracted from~\cite{Lee2018Concentrated} and contain $40000$ individual records with $58$ attributes including age, education level and so on.
The categorical attributes in the dataset are denoted by various integers.
The label of each sample describes whether the annual income of this individual is over $25$k.
In this model, the loss function is given by
\begin{equation}
F(\boldsymbol{w})=\frac{\kappa}{2}\Vert \boldsymbol{w}\Vert^{2}_{2}+\max\{y_{m}-\boldsymbol{w}^{\top}\mathcal{D}_{i, m}, 0\},
\end{equation}
where $\kappa>0$ is a regularization coefficient, $\mathcal{D}_{i,m}$ is the $m$-th sample in $\mathcal{D}_{i}$, $\mathcal{D}_{i}$ is the dataset of $i$-th MT and $y_{m}\in \{+1, -1\}$ for $m \in \{1,\ldots,\vert \mathcal{D}_{i}\vert\}$.

\emph{2)} MLP is trained using SGD consisting of single hidden layer with $256$ hidden units, where ReLU units and softmax of $10$ classes (corresponding to the $10$ digits) are applied.
We use the cross-entropy loss function and conduct experiments on the standard MNIST dataset for handwritten digit recognition consisting of $60000$ training examples and $10000$ testing examples~\cite{Lecun1998Gradient}.
Each example is a $28 \times 28$ size gray-level image of handwritten digits from $0$ to $10$.
In Fig.~\ref{fig:digits_visual_mlp}, we show several samples of the standard MNIST dataset with a visual illustration via MLP. The left figure and right figure in Fig.~\ref{fig:digits_visual_mlp} are derived from the original FL model and UDP model, respectively.
The visual results of MLP based FL with the interpretability technique are corresponding to the digit $0 \sim 9$ under the original FL and UDP based FL, respectively.
Basically speaking, these subfigures show that a digit understood by the machine model with the original FL are more distinct than the UDP based FL.

\emph{3)} CNN is trained using SGD consisting of single convolutional layer with the convolutional kernel size $5$ and the padding size $4$, where ReLU units and softmax of $10$ classes are applied.
We also use the cross-entropy loss function in this model.
The CIFA-$10$ dataset consists of $32 \times 32$ color images with three channels (RGB) in $10$ classes including ships, planes, dogs and cats.
Each class has $6000$ images where there are $40000$ examples for training, $10000$ for testing and $10000$ for validation.

Among them, the loss function for SVM is convex, whereas the loss function for MLP does not satisfy this condition.
The experimental results in this setting show that our theoretical results and proposed algorithm also work well for models (such as MLP) whose loss functions are not convex.

\textbf{Parameter Setting.} The total number of MTs $U$ in our experiments is set to $50$. In order to conduct our experiments conveniently, we consider the worst condition for model convergence with $\epsilon_{i} = \epsilon_{\text{p}}$ and $\delta_{i} = \delta_{\text{p}}$, $\forall i \in \mathcal{U}$, where $\epsilon_{\text{p}}$ and $\delta_{\text{p}}$ are the smallest values of permitting. In addition, we adopt a certain $\delta$ to study the effect of $\epsilon_{\text{p}}$ and set $\delta_{\text{p}}=0.001$. The threshold $\zeta$ in CRD method is set to $0.001$.
We can note that parameter clipping $C$ is a popular ingredient of SGD and ML for non-privacy reasons.
A proper value of clipping threshold $C$ should be considered for the DP based FL framework.
In the following experiments, we utilize the method in~[29] and choose $C$ by taking the median of the norms of the unclipped parameters over the course of training.
\begin{figure}[ht]
\centering
\includegraphics[width=2.85in,angle=0]{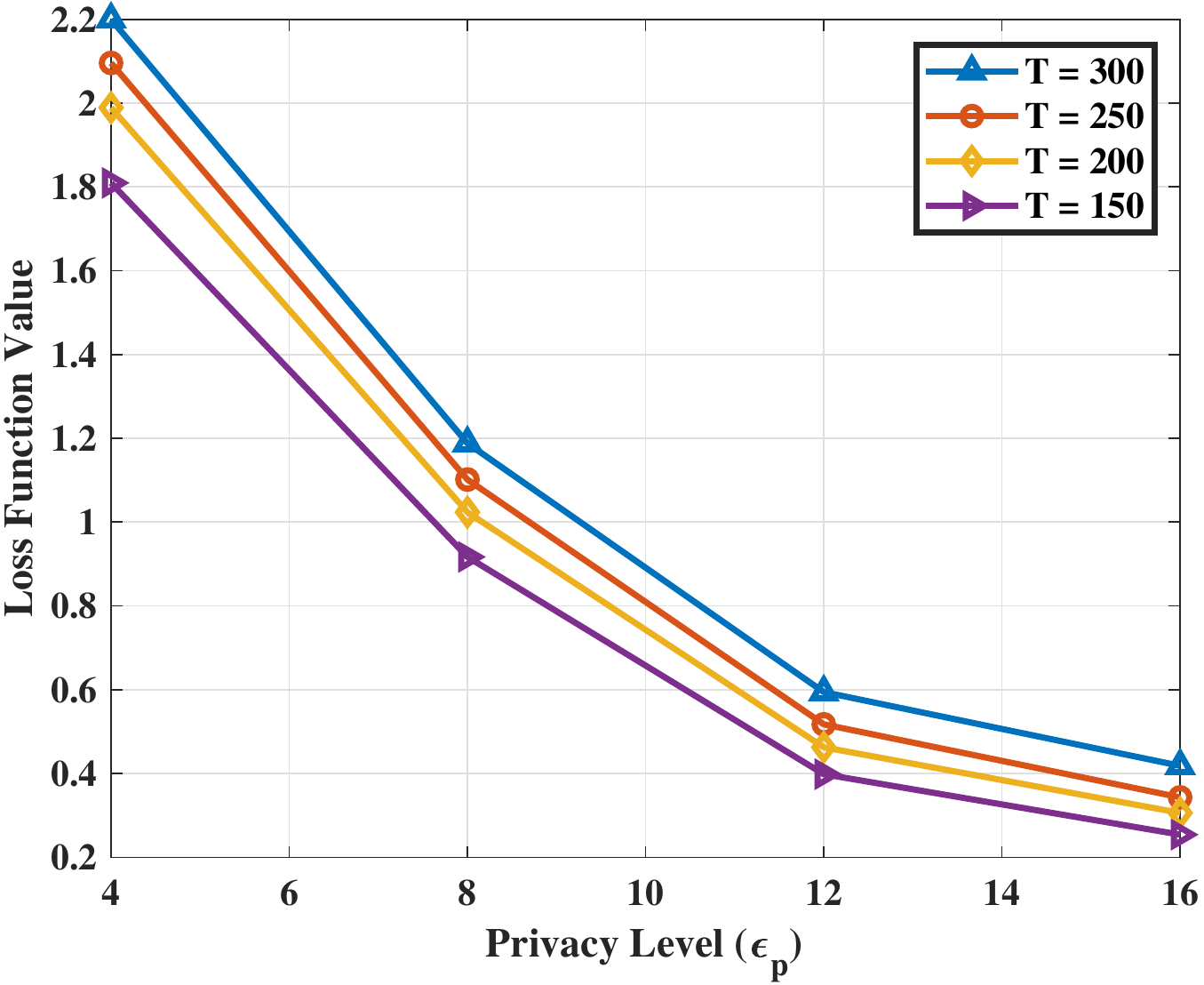}
\caption{Value of loss function using the UDP algorithm with CRD method ($\beta = 0.9$) with various initial values of $T$.}
\label{fig:InitialT_CRD}
\end{figure}
\begin{figure}[ht]
\centering
\includegraphics[width=2.85in,angle=0]{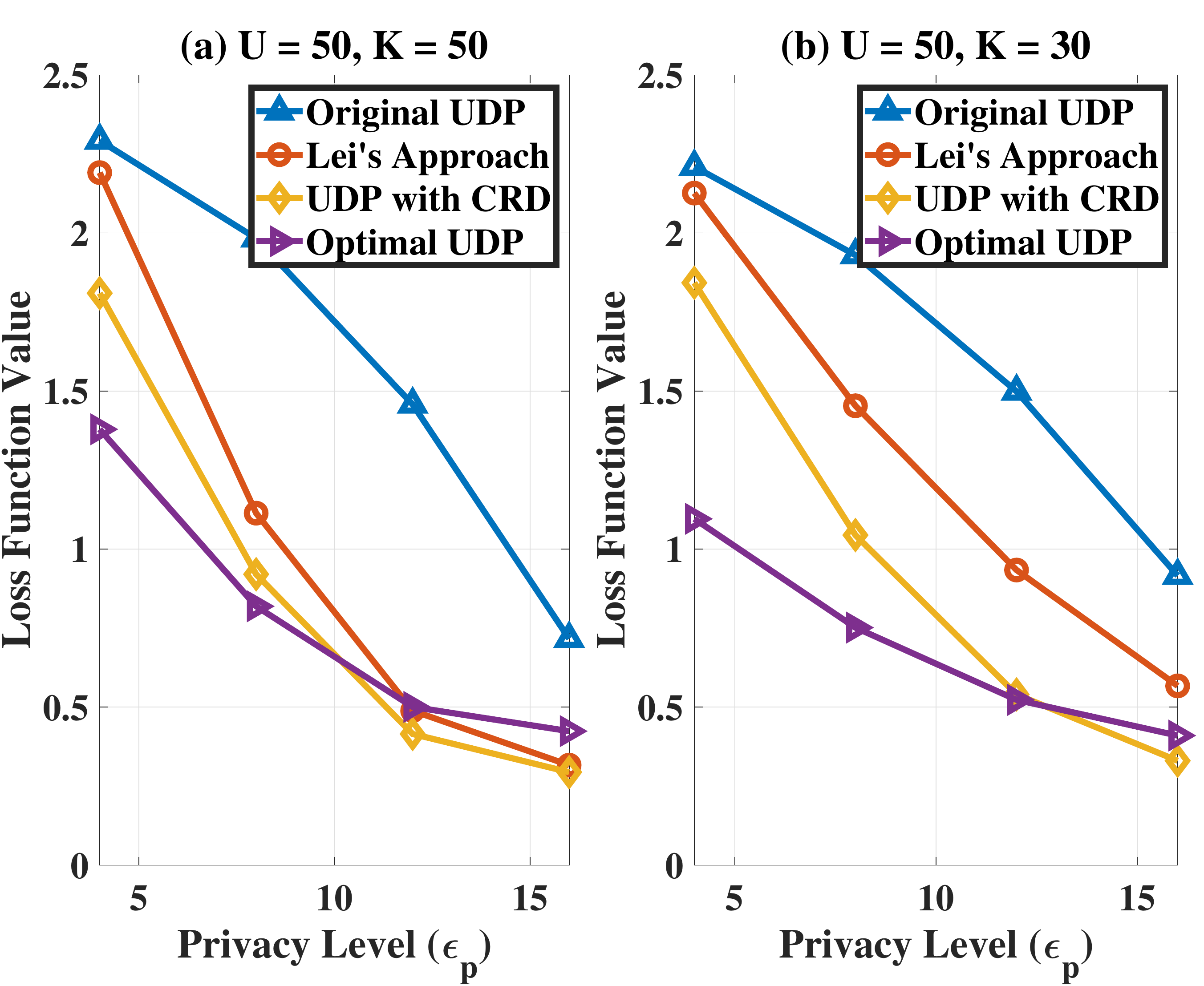}
\caption{Value of the loss function using the original UDP, Lei's approach, UDP with CRD method ($\beta = 0.9$) and the optimal UDP. (a) $U = 50, K=50$. (b) $U=50, K=30$.}
\label{fig:CRD_loss}
\end{figure}
\begin{figure}[ht]
\centering
\includegraphics[width=2.85in,angle=0]{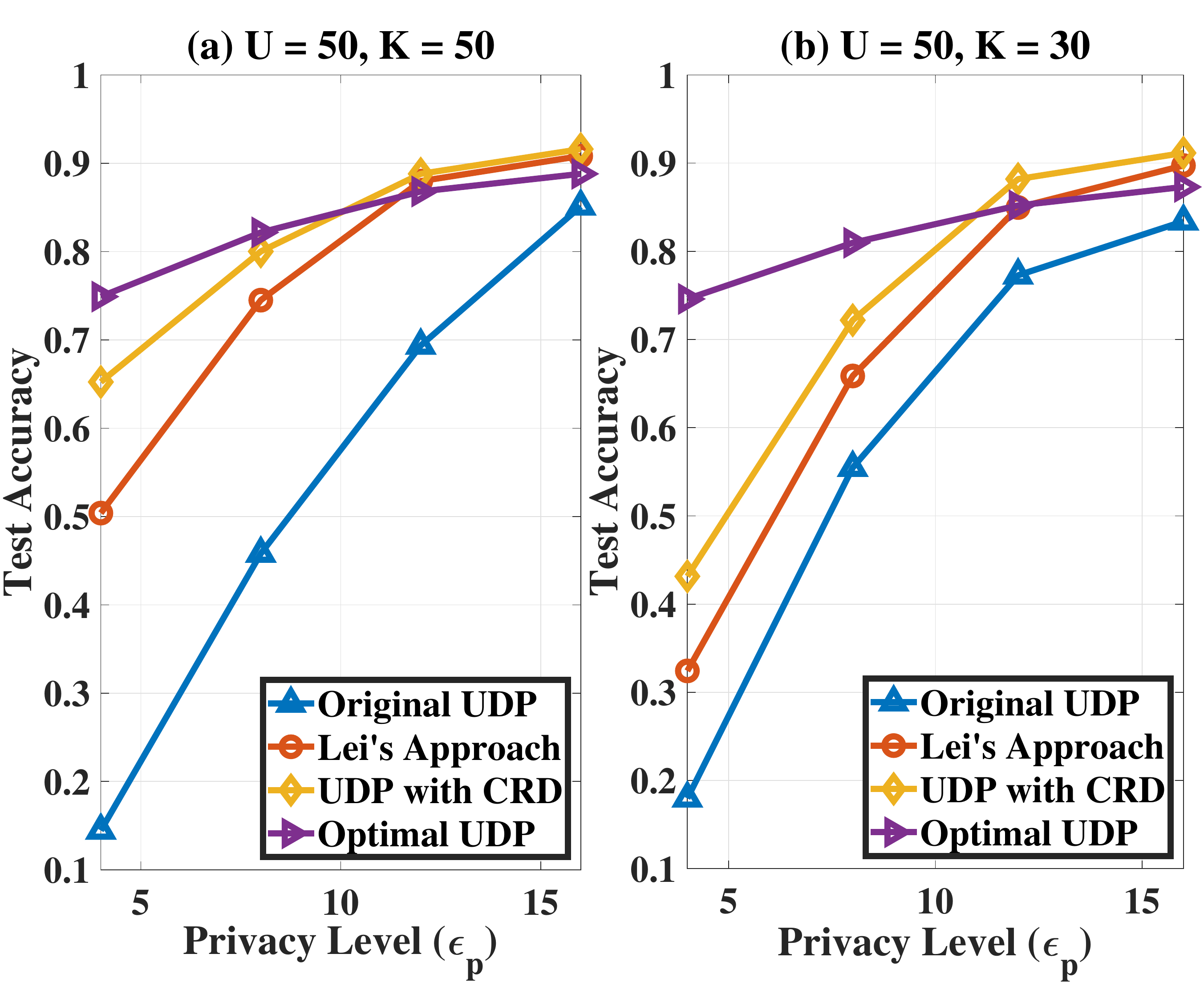}
\caption{Test accuracy using the original UDP, Lei's approach, UDP with CRD method ($\beta = 0.9$) and the optimal UDP. (a) $U = 50, K=50$. (b) $U=50, K=30$.}
\label{fig:CRD_acc}
\end{figure}
\begin{figure}[ht]
\centering
\includegraphics[width=2.85in,angle=0]{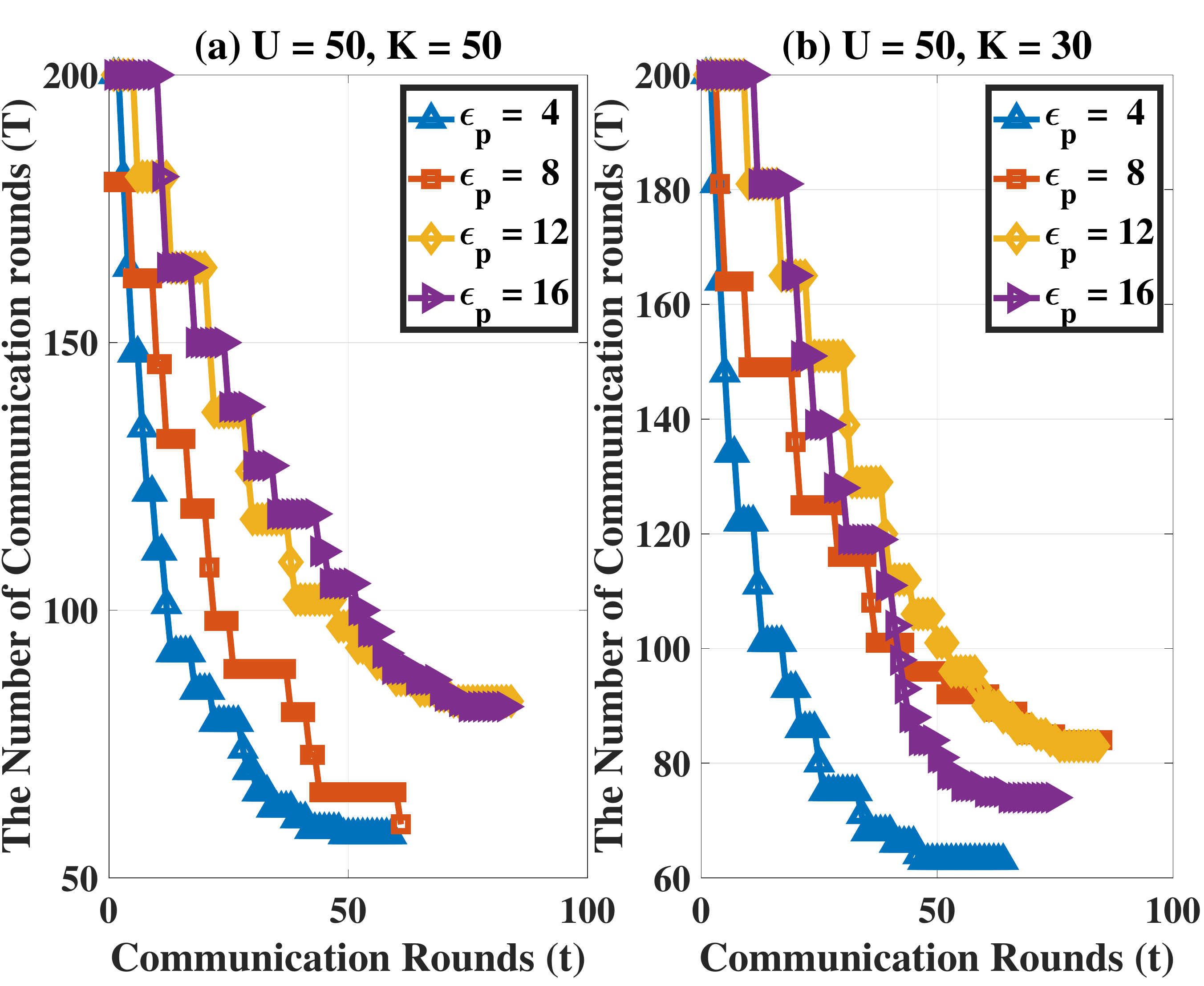}
\caption{The number of communication rounds using UDP algorithm with CRD method ($\beta = 0.9$). (a) $U = 50, K=50$. (b) $U=50, K=30$.}
\label{fig:CRD_MCR}
\end{figure}
\subsection{Effects of the number of Communication Rounds on UDP}
\begin{figure}[ht]
\centering
\includegraphics[width=2.85in,angle=0]{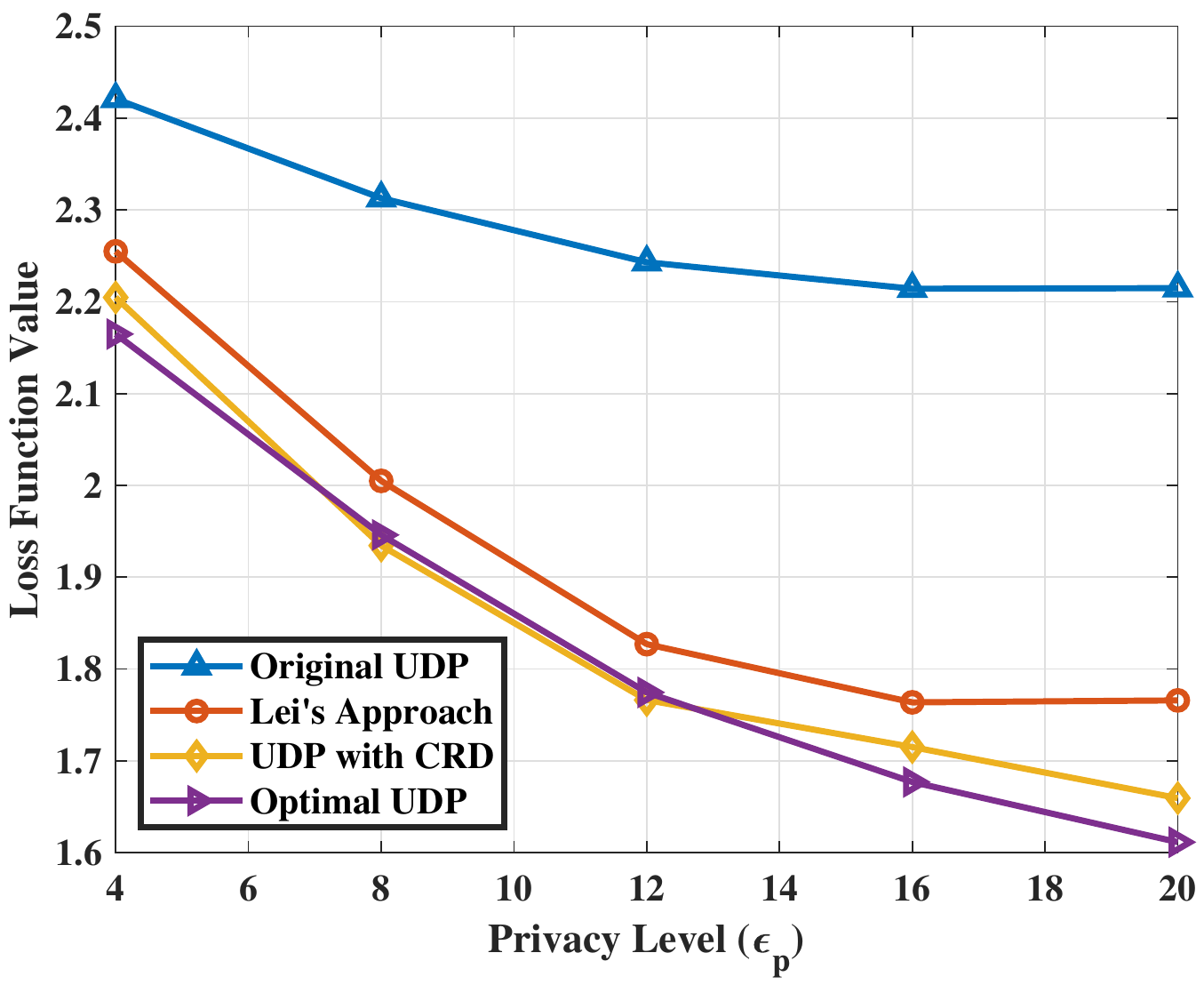}
\caption{Value of the loss function using the original UDP, Lei's approach, UDP with CRD method ($\beta = 0.9$) and the optimal UDP by a convolutional neural network (CNN) based FL for the multi-class classification task with the dataset CIFAR-10.}
\label{fig:CNN_cifar}
\end{figure}
\begin{figure}[ht]
\centering
\includegraphics[width=2.85in,angle=0]{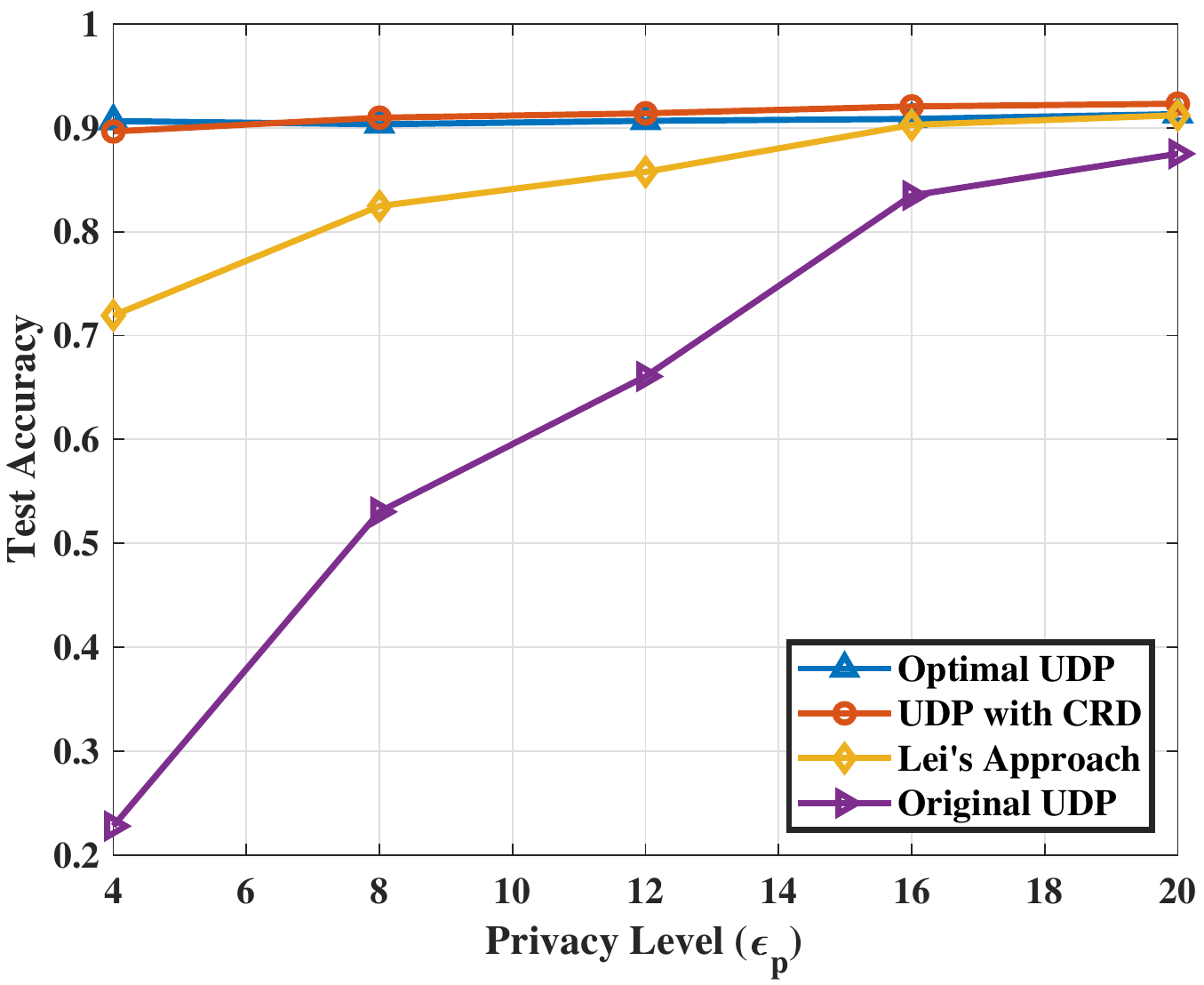}
\caption{Test accuracy using the original UDP, Lei's approach, UDP with CRD method ($\beta = 0.9$) and the optimal UDP under the non-IID data distribution.}
\label{fig:acc_data_noniid}
\end{figure}
\begin{figure}[ht]
\centering
\includegraphics[width=2.85in,angle=0]{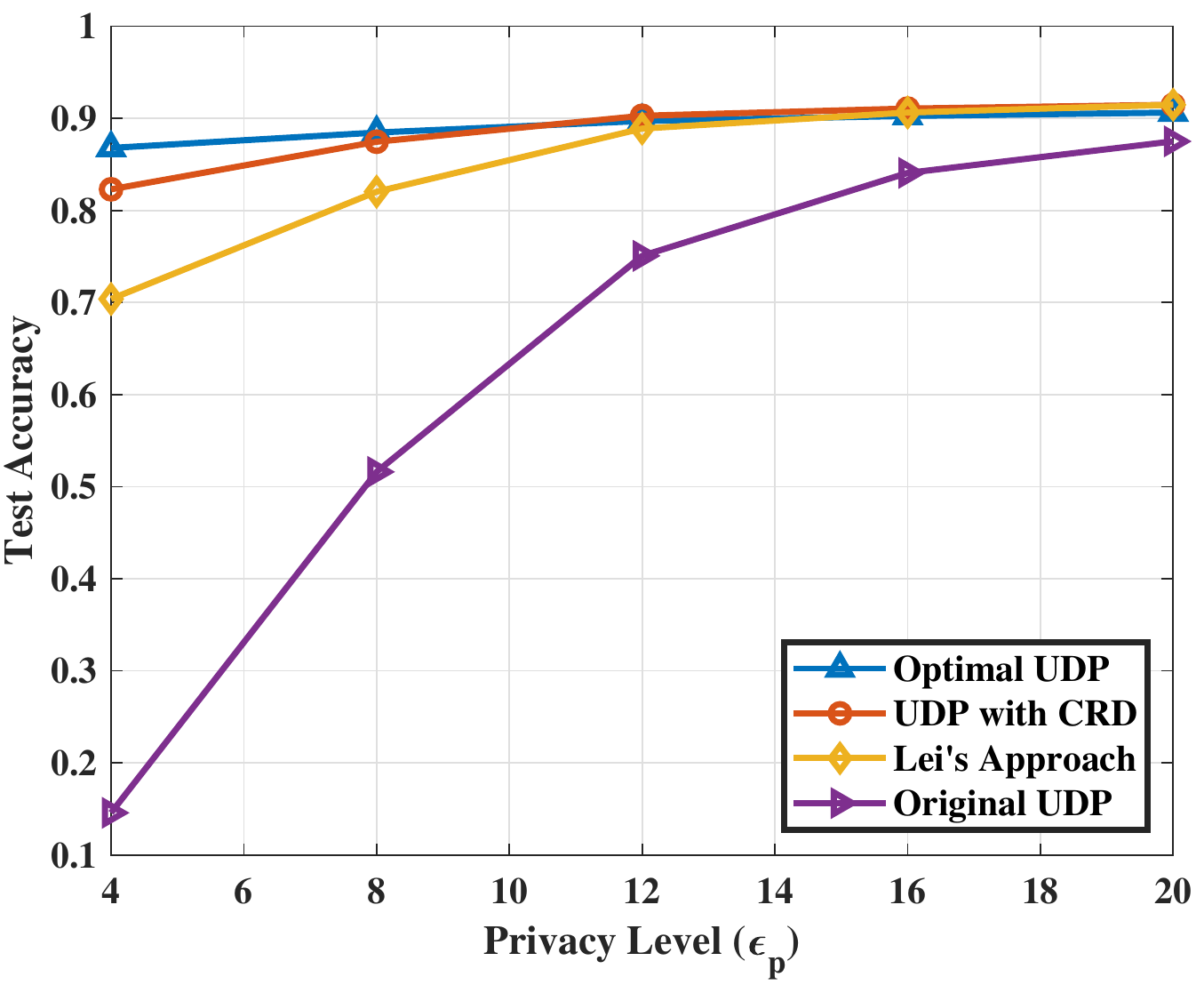}
\caption{Test accuracy using the original UDP, Lei's approach, UDP with CRD method ($\beta = 0.9$) and the optimal UDP under the unbalanced setting (different number of samples for different MTs).}
\label{fig:acc_data_distr}
\end{figure}
In this subsection, we verify the convex property of UDP for the value of $T$ with SVM and MLP models.
In Fig.~\ref{fig:SVM_ConvforNandK_T}, we show experimental results of testing loss as a function of $T$ with various privacy levels using SVM.
The size of local samples is set as $\vert\mathcal{D}_{i}\vert=128$.
The observation is in line with \emph{\textbf{Theorem~\ref{theorem:loss_ConvforK} and ~\ref{theorem:Covex_CR}}}, and the reason comes from the fact that a lower privacy level decreases the standard deviation of additive noises and the server can obtain better quality ML model parameters from MTs.
Fig.~\ref{fig:SVM_ConvforNandK_T} also implies that an optimal value of $T$ increases with a larger $\epsilon_{\text{p}}$.

Then, we show experimental results of the loss function value with respect to $T$ and privacy levels $\epsilon_{\text{p}}$ using the MLP network.
The size of local samples is set as $\vert\mathcal{D}_{i}\vert=800$ in this experiment.
Figs.~\ref{fig:ConvforN_T_eps_td} and~\ref{fig:ConvforK_T_eps_td} illustrate the expectation of the loss function, by varying privacy level $\epsilon_{\text{p}}$ and the value of $T$.
From Fig.~\ref{fig:ConvforN_T_eps_td}, we can observe that a large $T$ and a small $\epsilon_{\text{p}}$ may lead to a worse performance in the no sampling scenario.
As shown in the second scenario with sampling ratio $q=0.6$ in Fig.~\ref{fig:ConvforK_T_eps_td}, it also retains the same property.
\subsection{Effects of Parameters on CRD}
In this subsection, to evaluate our CRD algorithm, we apply the MLP model with the standard MNIST dataset.
Several experimental settings in our case study are introduced as follows.
The setting including the following main parts: 1) various initial values of $T$ using discounting method; 2) various privacy levels using CRD algorithm; and 3) various discounting factors using CRD algorithm.

We evaluate the effectiveness of our CRD method, and compare the results with the following benchmarks:
1) \emph{original UDP}, in which uniform privacy budget ($\epsilon_{\text{p}}$) allocation algorithm as well as the moments accountant method~\cite{Abadi2016Deep} is adopted in the UDP algorithm;
2) \emph{Lei's approach}~\cite{Yu2019Differentially} in which the STD of added noise will be reduced linearly until the privacy loss is lager than a preset $\epsilon_{\text{p}}$;
and 3) \emph{optimal UDP}, in which the UDP algorithm will be trained with the optimal $T$ (obtained by heuristic search).

\textbf{Initial Value of $T$.}
In the previous experiments, the initial value of $T$ is set as the default.
To examine the effect of initial $T$, we vary its value from $150$ to $300$ and measure the model convergence performance under  several fixed privacy levels.
We also choose this handwritten digit recognition task with the size of local samples $\vert\mathcal{D}_{i}\vert=800$ and the discounting factor $\beta = 0.9$.
Fig.~\ref{fig:InitialT_CRD} shows that when $T$ is closer to the optimal value of $T$ (the optimal value of $T$ by searching is shown in the above subsection), we can obtain a better convergence performance.
This observation is also in line with \emph{\textbf{Theorem~\ref{theorem:loss_ConvforK} and ~\ref{theorem:Covex_CR}}}.

\textbf{Privacy Level.}
We choose a handwritten digit recognition task with the initial number of communication rounds $T=200$ and the size of local samples $\vert\mathcal{D}_{i}\vert=800$.
We also set two different sampling ratio $q=1$ and $q=0.6$, which are corresponding to Fig.~\ref{fig:CRD_loss}, observed by (a) and (b), respectively.
In Fig.~\ref{fig:CRD_loss}, we describe how value of the loss function change with various values of the privacy level $\epsilon_{\text{p}}$ under original UDP (\textbf{Algorithm~1}), UDP with CRD (\textbf{Algorithm~2}) and optimal CRD (obtain the optimal $T$ by searching).
Fig.~\ref{fig:CRD_acc} shows the test accuracy corresponding to Fig.~\ref{fig:CRD_loss}.
We can note that using~\textbf{Algorithm~2} with discounting factor $\beta = 0.9$ can greatly improve the convergence performance in Fig.~\ref{fig:CRD_loss} (a) and (b), which is close to the optimal results.
Moreover, our UDP with CRD has a better performance than the optimal results with a large $\epsilon_{\text{p}}$, because the UDP algorithm with the optimal $T$ (obtained by searching) averagely allocates the privacy budget ($\epsilon_{\text{p}}$) during the training process but our algorithm can utilize the privacy budget more adaptively and efficiently.
In Fig.~\ref{fig:CRD_MCR}, we choose the same parameters with Fig.~\ref{fig:CRD_loss} and show the change of $T$ during training in one experiment under the UDP algorithm with CRD method ($\beta = 0.9$).
In Fig.~\ref{fig:CRD_MCR}, we can find that a smaller privacy level $\epsilon_{\text{p}}$ will have an early end in UDP algorithm with CRD method.
The intuition is that a larger $T$ can lead to a higher chance of information leakage and a larger noise STD of additive noises.
Then, the CRD method may be triggered and a decreased $T$ will be broadcasted to chosen MTs from the server.

In Fig.~\ref{fig:CNN_cifar}, we evaluate a CNN based federated learning for the multi-class classification task with the dataset CIFAR-10 in UDP with CRD, where each client has $800$ training samples locally.
The protection levels are set to $\epsilon=4$, $\epsilon=8$, $\epsilon=12$, $\epsilon=16$ and $\epsilon=20$ for this experiment. In addition, we set $N = 50$, $K = 30$, $T = 200$ and $\beta=0.9$. From Fig.~\ref{fig:CNN_cifar}, comparing with the original UDP and Lei's approach, we can note that using~\textbf{Algorithm~2} with discounting factor $\beta = 0.9$ can greatly improve the convergence performance.

\begin{figure}[ht]
\centering
\includegraphics[width=2.85in,angle=0]{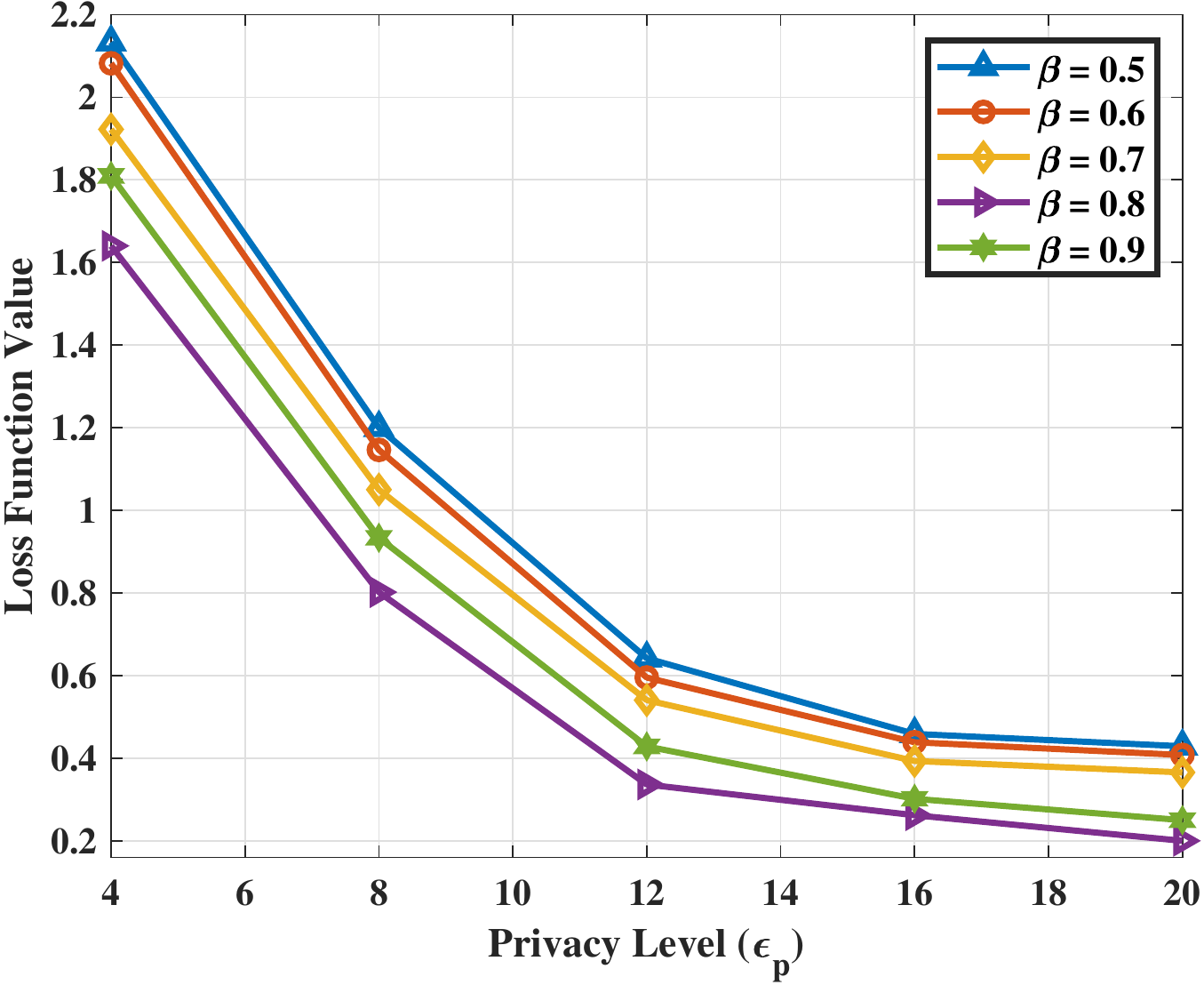}
\caption{Value of loss function using the UDP algorithm with CRD method with various discount factors.}
\label{fig:LinearDecay_loss}
\end{figure}
\begin{figure}[ht]
\centering
\includegraphics[width=2.85in,angle=0]{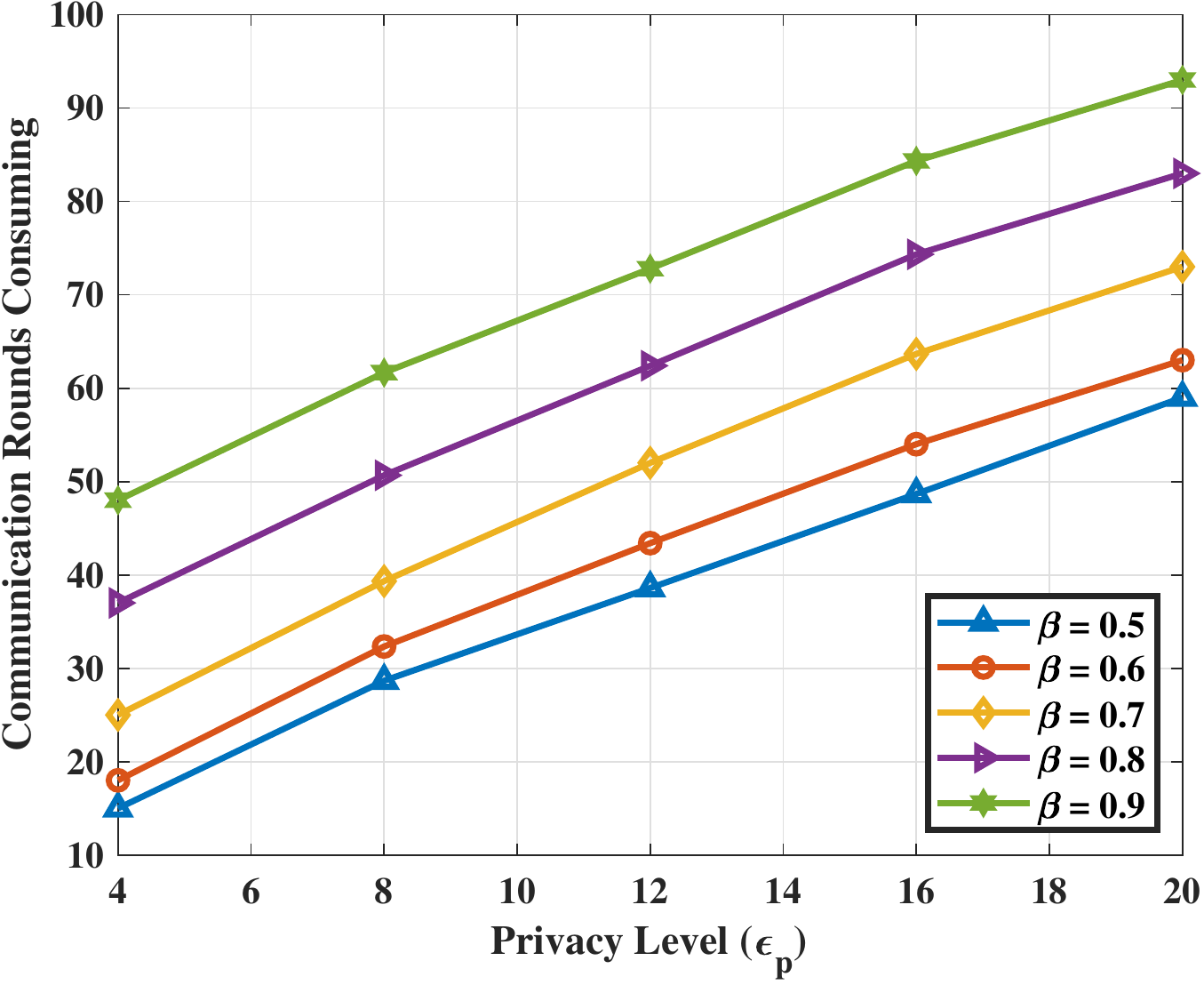}
\caption{Communication rounds consuming using the UDP algorithm with CRD method with various discount factors.}
\label{fig:LinearDecay_MCR}
\end{figure}

We also apply the MLP model with the standard MNIST dataset with the settings of non-IID data distribution and different number of samples (unbalanced) in UDP with CRD,
For the non-IID data distribution setting, each MT has four kinds of digits with the same amount, and the variety is different from all other MTs.
In the unbalanced setting, all MTs are divided into $5$ parts and the MT in each part has different number of training samples (400, 600, 800, 1000 and 1200, respectively ) locally.
The protection levels are set to $\epsilon=4$, $\epsilon=8$, $\epsilon=12$, $\epsilon=16$ and $\epsilon=20$ for this experiment. In addition, we set $N = K = 50$, $T = 200$ and $\beta=0.9$
From Figs.~\ref{fig:acc_data_noniid} and~\ref{fig:acc_data_distr}, we can note that using~\textbf{Algorithm~2} with discounting factor $\beta = 0.9$ can greatly improve the test accuracy comparing with the original UDP and Lei's approach.

\textbf{Discounting Factor.}
In Fig.~\ref{fig:LinearDecay_loss}, we vary the value of $\beta$ from $0.5$ to $0.9$ and plot their convergence results of the loss function.
The size of local samples and the initial number of communication rounds are set as $\vert\mathcal{D}_{i}\vert=800$ and $T=200$, respectively.
We observe that when the privacy level is fixed, a larger $\beta$ results in a slower decay speed of the $T$ which means careful adjustments of $T$ in the training and will benefit the convergence performance.
This is consistent with \emph{\textbf{Theorem~\ref{theorem:loss_ConvforK} and ~\ref{theorem:Covex_CR}}}.
With various values of different $\beta$, when we choose $\beta = 0.8$, the UDP algorithm with CRD method will have the best convergence performance.
In Fig.~\ref{fig:LinearDecay_MCR}, we show the communication rounds consuming (the number of required communication rounds) with various discounting factors using the same parameters with Fig.~\ref{fig:LinearDecay_loss}.
We find that more careful adjustments (corresponding to a larger $\beta$) will lead to more communication rounds consuming.
Hence, we can conclude that there is a tradeoff between communication rounds consuming and convergence performance by choosing $\beta$.
As a future work, it is of great interest to analytically evaluate the optimum value of $\beta$ to minimize the loss function.

\section{Conclusion}\label{Sec:Concl}
In this paper, we have introduced a UDP algorithm in FL to preserve MTs' privacy and proved that the UDP algorithm can satisfy the requirement of LDP under a certain privacy level by properly selecting the STD of additive noise processes.
Then, we have shown that there is an optimal number of communication rounds ($T$) in terms of convergence performance for a given protection level, which has motivated us to design an intelligent scheme for adaptively choosing the value of $T$.
To address this problem, we have proposed a CRD method for training an FL model, which will be triggered when the convergence performance stops improving.
Our experiments have demonstrated that this discounting method in the UDP algorithm can obtain a better tradeoff between convergence performance and privacy levels.
We have also noted that various initial values of $T$ and discounting factors $\beta$ bring out different convergence results.

It is noteworthy that the privacy budget allocation scheme greatly affects the quality of the FL training and the proposed CRD method also can be improved if we can obtain an exact FL performance prediction.
As a topic for future, it is of interest to design an effective privacy budget allocation scheme to improve the convergence performance with a given privacy level.


%
\ifCLASSOPTIONcaptionsoff
  \newpage
\fi



%

%
%

\bibliographystyle{IEEEtran}
\bibliography{reference}

\clearpage
\appendices
\section{Proof of Lemma~\ref{lemma:Comp_div}}\label{appendix:Comp_div}
Here, we want to compare the value between $D_{\nu_{1},\nu_{0}}$ and $D_{\nu_{0},\nu_{1}}$.
Hence, we conduct the property of $D_{\nu_{1},\nu_{0}}-D_{\nu_{0},\nu_{1}}$ and rewrite it as
\begin{equation}\label{equ:diff_div}
\begin{aligned}
D_{\nu_{1},\nu_{0}}&-D_{\nu_{0},\nu_{1}}=\int_{-\infty}^{+\infty} \nu_0\left(1-q+qe^{\frac{2z\Delta\ell-\Delta \ell^{2}}{2\sigma_{i}^{2}}}\right)^{\lambda+1} \mathrm{d}z\\
&-\int_{-\infty}^{+\infty} \nu_0\left(1-q+qe^{\frac{2z\Delta\ell-\Delta \ell^{2}}{2\sigma_{i}^{2}}}\right)^{-\lambda} \mathrm{d}z\\
&\overset{z=y+\frac{\Delta s}{2}}{=}\int_{-\infty}^{+\infty} e^{\frac{\left(y+\frac{\Delta \ell}{2}\right)^{2}}{2\sigma_{i}^{2}}}\left(1-q+qe^{\frac{y\Delta \ell}{\sigma_{i}^{2}}}\right)^{\lambda+1} \mathrm{d}z\\
&-\int_{-\infty}^{+\infty} e^{\frac{\left(y+\frac{\Delta \ell}{2}\right)^{2}}{2\sigma_{i}^{2}}}\left(1-q+qe^{\frac{y\Delta \ell}{\sigma_{i}^{2}}}\right)^{-\lambda} \mathrm{d}z.
\end{aligned}
\end{equation}
Transforming the negative part of this integral, we have
\begin{equation}\label{equ:neg_pos}
\begin{aligned}
D_{\nu_{1},\nu_{0}}&-D_{\nu_{0},\nu_{1}}=\int_{-\infty}^{0} e^{\frac{\left(y+\frac{\Delta\ell}{2}\right)^{2}}{2\sigma_{i}^{2}}}\left(1-q+qe^{\frac{y\Delta \ell}{\sigma_{i}^{2}}}\right)^{\lambda+1} \mathrm{d}z\\
&-\int_{-\infty}^{0} e^{\frac{\left(y+\frac{\Delta\ell}{2}\right)^{2}}{2\sigma_{i}^{2}}}\left(1+q\left(e^{\frac{y\Delta\ell}{\sigma_{i}^{2}}}-1\right)\right)^{-\lambda} \mathrm{d}z\\
&\overset{y=-z}{=}\int_{0}^{+\infty} e^{\frac{\left(z-\frac{\Delta\ell}{2}\right)^{2}}{2\sigma_{i}^{2}}}\left(1-q+qe^{\frac{-z\Delta\ell}{\sigma_{i}^{2}}}\right)^{\lambda+1} dz\\
&-\int_{0}^{+\infty} e^{\frac{\left(z-\frac{\Delta\ell}{2}\right)^{2}}{2\sigma_{i}^{2}}}\left(1-q+qe^{\frac{-z\Delta \ell}{\sigma_{i}^{2}}}\right)^{-\lambda} \mathrm{d}z.
\end{aligned}
\end{equation}
Hence, we can obtain that
\begin{equation}
\begin{aligned}
&D_{\nu_{1},\nu_{0}}-D_{\nu_{0},\nu_{1}}=\int_{0}^{+\infty}\left(\phi_{\text{positive}}(z)-\phi_{\text{negative}}(z)\right)\mathrm{d}z,
\end{aligned}
\end{equation}
where
\begin{equation}
\begin{aligned}
\phi_{\text{positive}}(z) &= e^{\frac{-\left(z+\frac{\Delta\ell}{2}\right)^{2}}{2\sigma_{i}^{2}}}\left(1-q+qe^{\frac{z\Delta\ell}{\sigma_{i}^{2}}}\right)^{\lambda+1}\\
&\quad+e^{\frac{-\left(z-\frac{\Delta\ell}{2}\right)^{2}}{2\sigma_{i}^{2}}}\left(1-q+qe^{\frac{-z\Delta\ell}{\sigma_{i}^{2}}}\right)^{\lambda+1}
\end{aligned}
\end{equation}
and
\begin{equation}
\begin{aligned}
\phi_{\text{negative}}(z) &= e^{\frac{-\left(z+\frac{\Delta\ell}{2}\right)^{2}}{2\sigma_{i}^{2}}}\left(1-q+qe^{\frac{z\Delta\ell}{\sigma_{i}^{2}}}\right)^{-\lambda}\\
&\quad+e^{\frac{-\left(z-\frac{\Delta\ell}{2}\right)^{2}}{2\sigma_{i}^{2}}}\left(1-q+qe^{\frac{-z\Delta\ell}{\sigma_{i}^{2}}}\right)^{-\lambda}.
\end{aligned}
\end{equation}
Then, let us use $\varphi(z)$ as
\begin{equation}
\varphi(z) \triangleq \frac{\phi_{\text{positive}}(z)}{\phi_{\text{negative}}(z)}.
\end{equation}
In order to develop the monotonicity of $\varphi(z)$, we define $ \theta = e^{\frac{z\Delta\ell}{\sigma_{i}^{2}}}$, and then have equation~\eqref{long_equ0}.
\begin{figure*}[ht]
\normalsize
\begin{equation}\label{long_equ0}
\begin{aligned}
\frac{\mathrm{d}\varphi(z)}{\mathrm{d}z}&=\frac{d\varphi(\theta)}{\mathrm{d}\theta}\frac{\mathrm{d}\theta(z)}{\mathrm{d}z}=\frac{\mathrm{d}\theta(z)}{\mathrm{d}z}\frac{1-q}{\phi^{2}_{\text{negative}}(z)}(1-q+q\theta)^{\lambda}\left(1-q+\frac{q}{\theta}\right)^{-\lambda-1}\left((\lambda+1)q\left(\theta-\frac{1}{\theta}\right)-(1-q+q\theta)\right)\\
&\quad+\frac{\mathrm{d}\theta(z)}{\mathrm{d}z}\frac{1-q}{\phi^{2}_{\text{negative}}(z)}\left(1-q+\frac{q}{\theta}\right)^{\lambda}\left(1-q+q\theta\right)^{-\lambda-1}\left(\lambda q\left(\theta-\frac{1}{\theta}\right)+(1-q+q\theta)\right).
\end{aligned}
\end{equation}
\vspace*{4pt}
\end{figure*}
Then, we define $ \gamma \triangleq(\lambda+1)q\left(\theta-\frac{1}{\theta}\right)/(1-q+q\theta)$ and we consider the condition that $\gamma<1$, we know
\begin{equation}
\frac{\gamma(1-q+q\theta)}{(\lambda+1)} =1-q+q\theta-\left(1-q+\frac{q}{\theta}\right).
\end{equation}
Therefore, we can obtain
\begin{equation}
\frac{1-q+\frac{q}{\theta}}{1-q+q\theta} = 1-\frac{\gamma}{\lambda+1}.
\end{equation}
And then, we can rewrite $\frac{\mathrm{d}\varphi(z)}{\mathrm{d}z}$ as equation~\eqref{long_equ1} on the top of the next page.
\begin{figure*}[ht]
\normalsize
\begin{equation}\label{long_equ1}
\begin{aligned}
\frac{\mathrm{d}\varphi(z)}{\mathrm{d}z}&=\frac{\mathrm{d}\theta(z)}{\mathrm{d}z}\frac{1-q}{\phi^{2}_{\text{negative}}(z)}(1-q+q\theta)^{\lambda+1}\left(1-q+\frac{q}{\theta}\right)^{-\lambda-1}\left(\gamma+1+\left(\frac{\lambda\gamma}{\lambda+1}+1\right)\left(1-\frac{\gamma}{\lambda+1}\right)^{2\lambda+1}\right).
\end{aligned}
\end{equation}
\hrulefill
\vspace*{4pt}
\end{figure*}
We define
\begin{equation}
\psi(\gamma)=\left(\gamma+1+\left(\frac{\lambda\gamma}{\lambda+1}+1\right)\left(1-\frac{\gamma}{\lambda+1}\right)^{2\lambda+1}\right).
\end{equation}
Then,
\begin{equation}
\frac{\mathrm{d}\psi(\gamma)}{\mathrm{d}\gamma}=1-\left(1+\frac{2\lambda\gamma}{\lambda+1}\right)\left(1-\frac{\gamma}{\lambda+1}\right)^{2\lambda},
\end{equation}
and
\begin{equation}
\frac{\mathrm{d}^{2}\psi(\gamma)}{\mathrm{d}\gamma^{2}}=\frac{2\lambda(2\lambda+1)\gamma}{(\lambda+1)^{2}}\left(1-\frac{\gamma}{\lambda+1}\right)^{2\lambda-1}\geq0.
\end{equation}
Because $\frac{\mathrm{d}\psi(\gamma)}{\mathrm{d}\gamma}|_{\gamma=0}=0$, we know $\frac{\mathrm{d}^{2}\psi(\gamma)}{\mathrm{d}\gamma^{2}}\geq 0$.
Considering $\varphi(0)=1$, we can conclude that $\varphi(z)\geq1$ and $D_{\nu_{1},\nu_{0}}\geq D_{\nu_{0},\nu_{1}}$.
This completes the proof. $\hfill\square$

\section{proof of Theorem~\ref{theorem:loss_ConvforK}}\label{appendix:ConvforK}
First, we know that the aggregated model by the server can be expressed as
\begin{equation}\label{equ:K_aggregation}
\boldsymbol{w}^{t+1} = \sum_{i \in \mathcal{K}}{p_{i}\boldsymbol{\widetilde{w}}_{i}^{t+1}}=\sum_{i \in \mathcal{K}}{p_{i}(\boldsymbol{w}_{i}^{t+1}}+\mathbf{n}^{t+1}_{i}).
\end{equation}
Then, we define
\begin{equation}
\mathbf{n}^{t+1} \triangleq \sum_{i \in \mathcal{K}}p_{i}\mathbf{n}^{t+1}_{i}.
\end{equation}
Using the second-order Taylor expansion, we have
\begin{equation}\label{equ:taylor_expan}
\begin{aligned}
F(\boldsymbol{w}^{t+1})-F(\boldsymbol{w}^{t})&\leq (\boldsymbol{w}^{t+1}-\boldsymbol{w}^{t})^{\top}\nabla F(\boldsymbol{w}^{t})\\
&\quad+\frac{L}{2}\Vert \boldsymbol{w}^{t+1}-\boldsymbol{w}^{t}\Vert^{2}.
\end{aligned}
\end{equation}
Because
\begin{equation}\label{equ:gd}
\boldsymbol{w}_{i}^{t+1}=\boldsymbol{w}^{t}-\eta \nabla F_{i}(\boldsymbol{w}^{t}),
\end{equation}
and then substitute inequation~\eqref{equ:gd} and inequation~\eqref{equ:K_aggregation} into inequation~\eqref{equ:taylor_expan}, we have
\begin{equation}\label{equ:long_equ2}
\begin{aligned}
F(\boldsymbol{w}^{t+1})&- F(\boldsymbol{w}^{t})\leq \left(\mathbf{n}^{t+1}-\eta\sum_{i\in \mathcal{K}}{p_{i}\nabla F_{i}(\boldsymbol{w}^{t})}\right)^{\top}\nabla F(\boldsymbol{w}^{t})\\
&\quad+\frac{L}{2}\left\Vert \sum_{i\in \mathcal{K}}{p_{i}(\mathbf{n}_{i}^{t+1}-\eta\nabla F_{i}(\boldsymbol{w}^{t}))}\right\Vert^{2}\\
&=\frac{\eta^2L}{2}\left\Vert \sum_{i\in \mathcal{K}}{p_{i}\nabla F_{i}(\boldsymbol{w}^{t})}\right\Vert^{2}+\frac{L}{2}\left\Vert\sum_{i\in \mathcal{K}}p_{i}\mathbf{n}_{i}^{t+1}\right\Vert^2\\
&\quad-\eta\nabla F(\boldsymbol{w}^{t})^{\top}\sum_{i\in \mathcal{K}}{p_{i}\nabla F_{i}(\boldsymbol{w}^{t})}.
\end{aligned}
\end{equation}
The expected objective function $F(\boldsymbol{w}^{t+1})$ can be expressed as
\begin{multline}\label{equ:exp_obj}
\mathbb{E}\{F(\boldsymbol{w}^{t+1})\}\leq F(\boldsymbol{w}^{t})-\eta\Vert \nabla F(\boldsymbol{w}^{t})\Vert^{2}\\
+\frac{\eta^2L}{2}\mathbb{E}\{\Vert \sum_{i\in \mathcal{K}}{p_{i}\nabla F_{i}(\boldsymbol{w}^{t})}\Vert^{2}\}
+\frac{L}{2}\mathbb{E}\{\Vert\mathbf{n}^{t+1}\Vert^2\}.
\end{multline}
With an assumption that $p_{i} = 1/K$ and we have
\begin{equation}\label{equ:ConvforK_0}
\begin{aligned}
&\mathbb{E}\left\{\left\Vert \sum_{i\in \mathcal{K}}{p_{i}\nabla F_{i}(\boldsymbol{w}^{t})}\right\Vert^{2}\right\}= \frac{1}{UK}\sum_{i\in \mathcal{U}}{\Vert \nabla F_{i}(\boldsymbol{w}^{t})\Vert^{2}}\\
&\quad\quad+\frac{K-1}{UK(U-1)}\sum_{i\in \mathcal{U}}\sum_{j\in \mathcal{U}/i}{\left[\nabla F_{i}(\boldsymbol{w}^{t})\right]^{\top}\nabla F_{j}(\boldsymbol{w}^{t})}\\
&\quad=\left(\frac{1}{UK}-\frac{K-1}{UK(U-1)}\right)\sum_{i\in \mathcal{U}}{\Vert \nabla F_{i}(\boldsymbol{w}^{t})\Vert^{2}}\\
&\quad\quad+\frac{K-1}{UK(U-1)}\left(\sum_{i\in \mathcal{U}}\nabla F_{i}(\boldsymbol{w}^{t})\right)^2\\
&\quad= \frac{U(K-1)}{K(U-1)}\Vert\nabla F(\boldsymbol{w}^{t})\Vert^2\\
&\quad\quad+\frac{U-K}{UK(U-1)}\sum_{i\in \mathcal{U}}{\Vert \nabla F_{i}(\boldsymbol{w}^{t})\Vert^{2}}.
\end{aligned}
\end{equation}
According to~\textbf{Assumption~\ref{assu:conv_FL}}, we know
\begin{multline}
\mathbb{E}\{\varepsilon_{i}\}=\frac{1}{U}\sum_{i\in \mathcal{U}}\varepsilon_{i}=\frac{1}{U}\sum_{i\in \mathcal{U}}{\Vert \nabla F_{i}(\boldsymbol{w}^{t})-\nabla F(\boldsymbol{w}^{t})\Vert^{2}}\\
=\frac{1}{U}\sum_{i\in \mathcal{U}}{\Vert \nabla F_{i}(\boldsymbol{w}^{t})\Vert^{2}}-\Vert\nabla F(\boldsymbol{w}^{t})\Vert^2,
\end{multline}
Subtracting $\mathbb{E}\{\varepsilon_{i}\}$ into inequation~\eqref{equ:ConvforK_0}, we have
\begin{equation}\label{equ:upper_bound}
\begin{aligned}
&\mathbb{E}\left\{\left\Vert \sum_{i\in \mathcal{K}}{p_{i}^{\mathcal{K}}\nabla F_{i}(\boldsymbol{w}^{t})}\right\Vert^{2}\right\}\\
&\quad=\frac{U-K}{UK(U-1)}\sum_{i\in \mathcal{U}}{\Vert \nabla F_{i}(\boldsymbol{w}^{t})-\nabla F(\boldsymbol{w}^{t})\Vert^{2}}\\
&\quad\quad+\Vert\nabla F(\boldsymbol{w}^{t})\Vert^2
\leq\frac{(U-K)\varepsilon}{K(U-1)}+\Vert\nabla F(\boldsymbol{w}^{t})\Vert^2.
\end{aligned}
\end{equation}
Then, subtracting inequation~\eqref{equ:upper_bound} into inequation~\eqref{equ:exp_obj}, we have
\begin{multline}\label{Appendix_D_1}
\mathbb{E}\{F(\boldsymbol{w}^{t+1})\}\leq F(\boldsymbol{w}^{t})-\eta\left(\frac{\eta L}{2}-1\right)\Vert \nabla F(\boldsymbol{w}^{t})\Vert^{2}\\
+\frac{L}{2}\mathbb{E}\{\Vert\mathbf{n}^{t+1}\Vert^{2}\}+\frac{(U-K)\varepsilon}{K(U-1)}\\
\leq +\eta\left(\frac{\eta L}{2}-1\right)\Vert \nabla F(\boldsymbol{w}^{t})\Vert^{2}\\
+F(\boldsymbol{w}^{t})+\frac{L}{2}\mathbb{E}\{\Vert\mathbf{n}^{t+1}\Vert^{2}\}+\frac{\eta^{2} L(U-K)\varepsilon}{2K(U-1)}.
\end{multline}
Subtracting $F(\boldsymbol{w}^{*})$ into both sides of inequation~\eqref{Appendix_D_1}, we have
\begin{equation}\label{Appendix_D_2}
\begin{aligned}
\mathbb{E}\{F(\boldsymbol{w}^{t+1})\}&-F(\boldsymbol{w}^{*})\leq \mathbb{E}\{F(\boldsymbol{w}^{t})\}-F(\boldsymbol{w}^{*})\\
&+\frac{\eta^{2} L(U-K)\varepsilon}{2K(U-1)}+\eta\left(\frac{\eta L}{2}-1\right)\Vert \nabla F(\boldsymbol{w}^{t})\Vert^{2}\\
&+\frac{L}{2}\mathbb{E}\{\Vert\mathbf{n}^{t+1}\Vert^{2}\}.
\end{aligned}
\end{equation}
Considering Polyak-Lojasiewicz condition and applying inequation~\eqref{Appendix_D_2} recursively, and then considering the independence of additive noises, we know
\begin{multline}
\mathbb{E}\{F(\boldsymbol{w}^{t})\}-F(\boldsymbol{w}^{*})\leq \left(1-2\mu\eta+\mu\eta^{2}L\right)^{t}(F(\boldsymbol{w}^{0})-F(\boldsymbol{w}^{*}))\\
+\frac{L^2(1-(1-2\mu\eta+\mu\eta^{2}L)^{t})}{2\mu}\bigg{(}\mathbb{E}\{\Vert\mathbf{n}\Vert^{2}\}\\
+\frac{\eta^{2} (U-K)\varepsilon}{K(U-1)}\bigg{)}.
\end{multline}
Substituting inequation~\eqref{equ:DP_ConvforK} into the above inequality, we have
\begin{multline}
\mathbb{E}\{F(\boldsymbol{w}^{t})\}-F(\boldsymbol{w}^{*})\leq \left(1-2\mu\eta+\mu\eta^{2}L\right)^{t}(F(\boldsymbol{w}^{0})-F(\boldsymbol{w}^{*}))\\
+\frac{L(1-(1-2\mu\eta+\mu\eta^{2}L)^{t})}{\mu}\left(\frac{L\Delta\ell^{2}qt}{U}\sum_{i=1}^{U}\frac{\ln(1/\delta_{i})}{\epsilon_{i}^{2}}\right.\\
\left.+\frac{\eta^{2} L(U-K)\varepsilon}{2K(U-1)}\right).
\end{multline}
Hence, the convergence bound can be given as
\begin{multline}\label{equ:ConvforK_loss}
\mathbb{E}\{F(\boldsymbol{w}^{T})\}-F(\boldsymbol{w}^{*})\leq A^{T}(F(\boldsymbol{w}^{0})-F(\boldsymbol{w}^{*}))\\
+(1-A^T)\left(\frac{\kappa_{0}TK}{U^2}\sum_{i=1}^{U}\frac{\ln(1/\delta_{i})}{\epsilon_{i}^{2}}+\frac{\kappa_{1}U(U-K)}{K(U-1)}\right),
\end{multline}
where $A = 1-2\mu\eta+\mu\eta^{2}L$, $\kappa_{0} = \frac{L^2\Delta\ell^2}{\mu}$ and $\kappa_{1} = \frac{\eta^{2} L^{2}\varepsilon}{2\mu}$.
This completes the proof. $\hfill\square$

\end{document}